\documentclass[journal]{IEEEtran}
\usepackage{amsmath,amssymb,amsfonts}
\usepackage{algorithmic}
\usepackage{algorithm}
\usepackage{array}
\usepackage[caption=false,font=normalsize,labelfont=sf,textfont=sf]{subfig}
\usepackage{textcomp}
\usepackage{color}
\usepackage{stfloats}
\fnbelowfloat
\usepackage{cuted}
\usepackage{capt-of}
\usepackage{url}
\usepackage{verbatim}
\usepackage{graphicx}
\usepackage{enumitem}
\usepackage{makecell}
\usepackage{multirow}
\usepackage{booktabs}
\usepackage{placeins}
\usepackage{etoolbox}
\usepackage{amsthm}
\usepackage{units}
\newtheoremstyle{upright}
{3pt}
{3pt}
{\normalfont}
{}
{\bfseries}
{.}
{ }
{}

\theoremstyle{upright}

\makeatletter
\g@addto@macro\normalsize{%
\setlength{\abovedisplayskip}{5pt plus 2pt minus 1pt}
\setlength{\belowdisplayskip}{5pt plus 2pt minus 1pt}
\setlength{\belowdisplayshortskip}{4pt plus 2pt minus 1pt}}
\makeatother
\begin{document}
\flushbottom

\title{ISB-NMPC: Interception-Oriented\\Spherical-Bearing
Nonlinear Model Predictive Control for Quadrotors}

\author{\footnotesize Li Hong\textsuperscript{1},
Linkai Liu\textsuperscript{2},
Xiangyu Ren\textsuperscript{2},
Shuli Lv\textsuperscript{2},
Chen Min\textsuperscript{1},
Chunjiang Wu\textsuperscript{3},
Yujun Zhang\textsuperscript{1},
Dingchen Zeng\textsuperscript{2},

and
Quan Quan\textsuperscript{2,*}%
\thanks{*Corresponding author: Quan Quan (e-mail: qq\_buaa@buaa.edu.cn).}%
\thanks{\textsuperscript{1}Li Hong, Chen Min, and Yujun Zhang are with Shen Yuan Honors College, Beihang University, Beijing, China (e-mail: hongli@buaa.edu.cn for Li Hong).}%
\thanks{\textsuperscript{2}Linkai Liu, Xiangyu Ren, Shuli Lv, Dingchen Zeng, and Quan Quan are with the School of Automation Science and Electrical Engineering, Beihang University, Beijing, China.}%
\thanks{\textsuperscript{3}Chunjiang Wu is with the Hangzhou International Innovation Institute, Beihang University, Hangzhou, China.}%
}

\maketitle

\begin{abstract}

Quadrotor interceptors equipped with strapdown cameras offer a low-cost
solution for physically countering unauthorized UAVs. However, this
configuration inherently couples vehicle maneuvering and camera pointing,
creating a conflict between rapid closing and maintaining target visibility.
We present interception-oriented spherical-bearing nonlinear
model predictive control (ISB-NMPC), which couples spherical-bearing prediction
with nonlinear quadrotor dynamics to jointly optimize target-directed closing
motion and visibility. An attitude-compensated extended Kalman filter on
$\mathbb S^2$ estimates target bearing motion, compensates for perception
delay, and predicts target bearings to guide scale-adaptive detection.
Across 30 paired simulation trials, ISB-NMPC achieves an interception success
rate of $86.7\%$, the highest among the evaluated controllers, with a mean per-trial target-visible time of $99.11\%$.

Outdoor flights demonstrate acquisition of a
$30\,\mathrm{cm}$ target at approximately $150\,\mathrm{m}$, with interceptor
speed reaching $17\,\mathrm{m\,s^{-1}}$ in the long-range flight and
approximately $14\,\mathrm{m\,s^{-1}}$ while pursuing a quadrotor-towed
maneuvering target.
The software and hardware are open source at
\url{https://github.com/HOLYNOTBAD/ISB-NMPC} and
\url{https://github.com/HOLYNOTBAD/holydrone}, respectively.

\end{abstract}


\section{Introduction}

\begin{figure}[!t]
    \centering
    \captionsetup[subfloat]{font=footnotesize,labelfont=rm,textfont=rm}
    \subfloat[Long-range interception of a 30~cm quasi-static target over an
    approximately 150~m approach.
    \label{fig:flight_validation}]{%
        \includegraphics[width=\columnwidth]{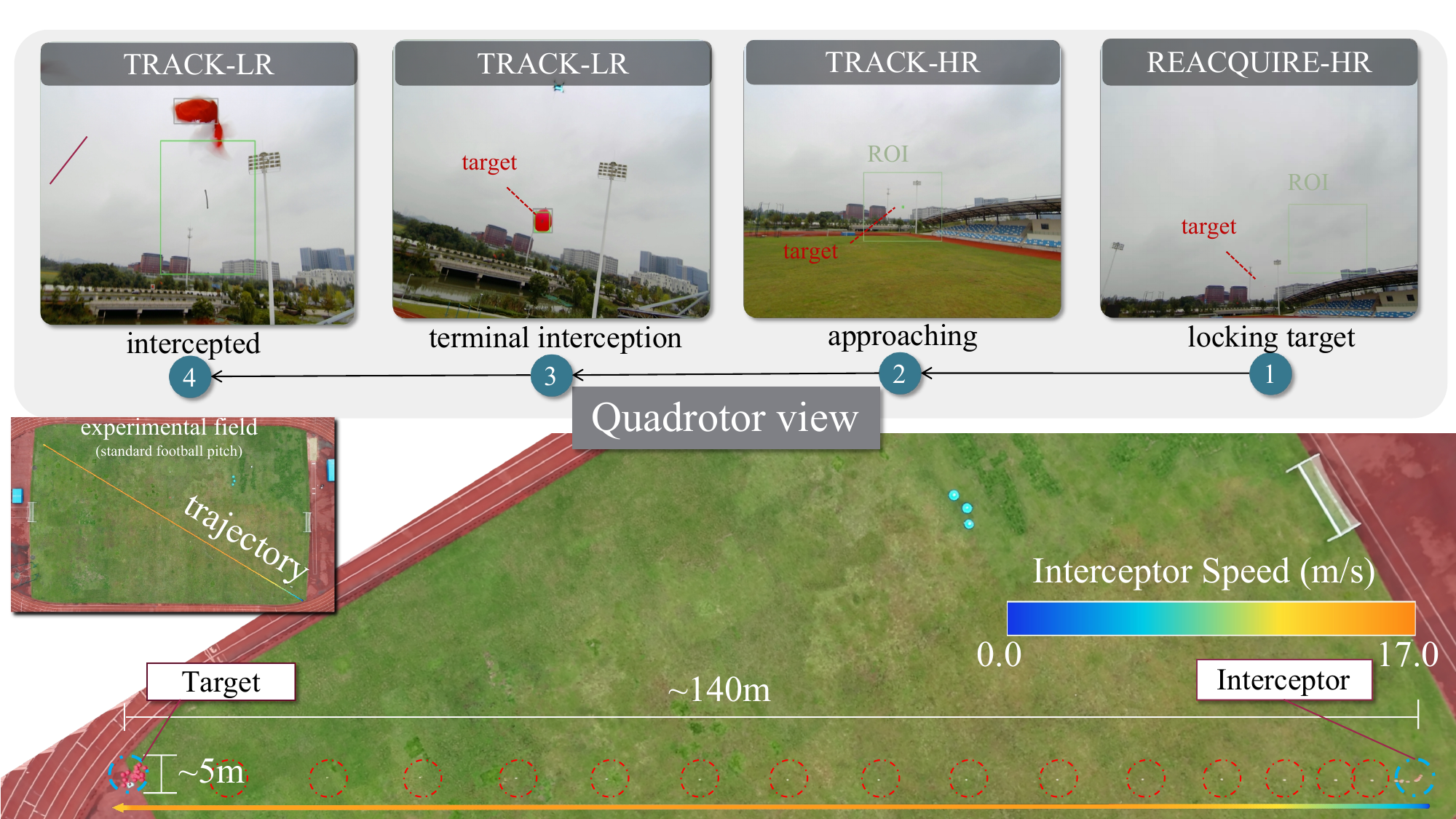}}\\[0.6ex]
    \subfloat[Pursuit--evasion interception of a 30~cm-diameter
    quadrotor-towed target from an initial separation of approximately 50~m.
    \label{fig:flight_dynamic_validation}]{%
        \includegraphics[width=\columnwidth]{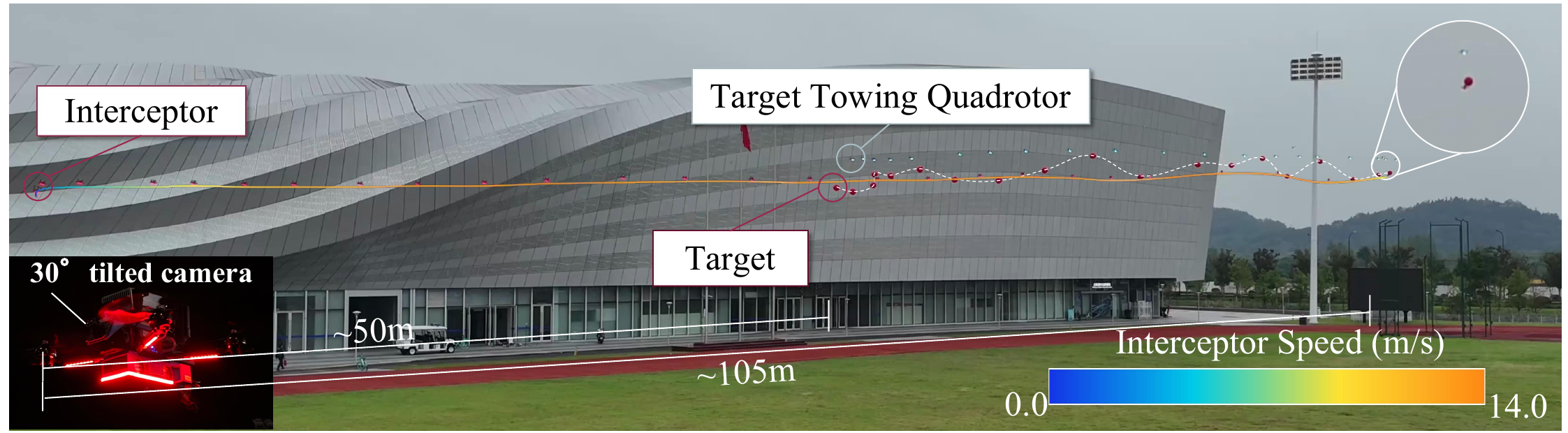}}
    \caption{
    Outdoor flight experiments with ISB-NMPC.
    Trajectories are colored by interceptor speed; insets show onboard views.}
    \label{fig:framework_overview}
\end{figure}

The growth of low-altitude aerial applications makes shared-airspace safety
increasingly important. Non-cooperative small targets create a practical need
for affordable onboard interception
systems~\cite{liBearingOnlyFollowing2023,pliskaSafeMidairDrone2024}.
A strapdown monocular camera offers lightweight, mechanically simple sensing,
but the interceptor must acquire the target, retain visual feedback, and
close to a capture region without cooperative target-state measurements.
Figure~\ref{fig:framework_overview} illustrates the long-range and
maneuvering-target scenarios considered here.

\begin{figure}[!tbp]
\centering
\includegraphics[width=\columnwidth]{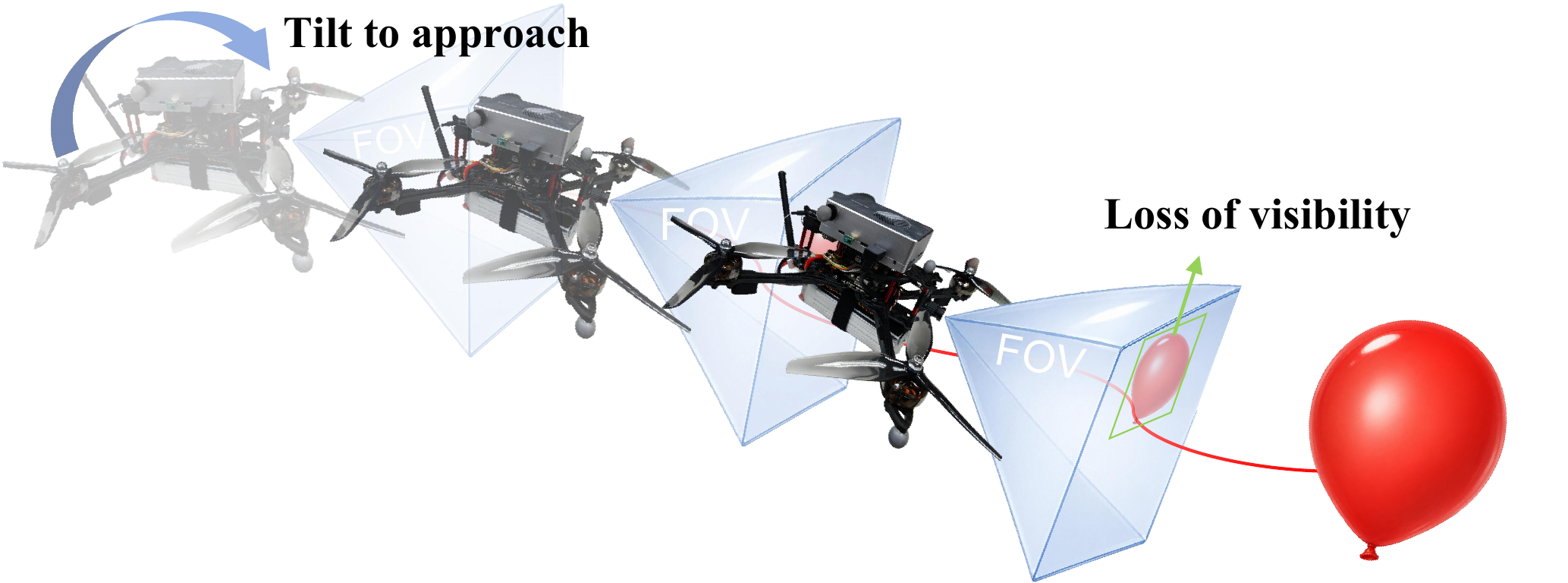}
\caption{Closing--visibility trade-off: target-directed
acceleration tilts the strapdown camera and may reduce the target's FOV
margin.}
\label{fig:interception_process}
\end{figure}

This mounting arrangement creates a fundamental control conflict.
Target-directed acceleration requires tilting the thrust vector, which rotates
the camera and can push the target toward the field-of-view (FOV) boundary.
Recentering the target can, in turn, restrict closing acceleration.
Thus, the same thrust and attitude motion govern both closing efficiency and
visibility~\cite{bourquardezImagebasedVisualServo2009},
requiring coordination under nonlinear dynamics and actuator limits
(Fig.~\ref{fig:interception_process}). Camera rotation, detection delay, and
target-scale variation further affect the supporting visual feedback.

Existing approaches follow three complementary routes: target-state
estimation and trajectory planning, visual servoing and line-of-sight (LOS)
guidance, and
perception-aware model predictive control (MPC). The first plans and tracks
interception trajectories.
Optimization-based planning explicitly handles engagement geometry, but its
execution depends on the available target-state
information~\cite{pliskaSafeMidairDrone2024}.
A second route generates motion directly from visual feedback or relative
geometry. Image-based visual servoing (IBVS) regulates image features, whereas
position-based visual servoing (PBVS) uses estimated relative
pose~\cite{chaumetteVisualServoControl2006,xuVisualServoReview2026};
barrier-function constraints can preserve
visibility~\cite{zhengVisibilityGuaranteed2019}.

For interception, LOS guidance connects visual feedback to closing.
Yang et al. developed high-speed image-based
control~\cite{yangHighSpeedInterceptionMulticopter2025a} and analyzed
LOS-constrained multirotor
interceptability~\cite{yangLineofSightConstrainedMulticopterInterceptability2025a}.
Related studies address planar-sector
guidance~\cite{liuPlanarsectorGuidanceInterception}, pursuit--evasion
interceptability~\cite{zhangInterceptabilityVisionguidedUAVs2025}, and persistent
interception~\cite{yangPersistentInterception2026}. Bearing-only following also
uses helical motion to improve observability while maintaining a prescribed
geometry~\cite{liBearingOnlyFollowing2023}. These methods establish useful
feedback structures; rapid capture additionally requires coordinating closing
demand with visibility-preserving vehicle motion.

The third coordinates motion and sensing over a prediction horizon.
Falanga et al.~\cite{falangaPAMPCPerceptionawareModel2018} incorporate perception
into quadrotor motion optimization. Related formulations address reactive
collision and occlusion avoidance~\cite{peninVisionBasedReactivePlanning2018}
and combine visibility with actuator
constraints~\cite{dmytrukPerceptionAwareNMPC2022}, while robust visual
predictive controllers address tracking and visibility
constraints~\cite{zhangRobustVisualNMPC2021}.
Qin et al.~\cite{qinPerceptionawareImagebasedVisual2023} combine quaternion-based
spherical features and visibility constraints for aggressive flight, and
Li et al.~\cite{liRobocentricVisualServo2023} integrate estimation and control in
a robocentric model.
MPC has also been applied to non-cooperative aerial interception and
impact-angle-constrained visual
engagement~\cite{srivastavaAerialInterceptionMPC2022,cetinVisualInterceptionMPC2023}.

For strapdown interception, the remaining question is how to choose
closing direction and speed under visibility and actuation constraints.
Uncoordinated closing can lose visual feedback, whereas recentering alone can
delay capture. We connect LOS guidance to spherical prediction through an
optimized closing-speed state; paired simulations and flight observations
examine closing and visual retention together.

The supporting perception front end draws on depth-independent
bearing estimation~\cite{hamelVisualServoingUnderactuated2002,mebarkiNonlinearVisualControl2015},
small-object slicing~\cite{akyonSlicingAided2022}, and delayed-feedback
estimation~\cite{guoDelayedVisualLanding2024} to address camera rotation,
target-scale variation, and latency.

We propose interception-oriented spherical-bearing nonlinear model
predictive control (ISB-NMPC), predicting $SO(3)$-lifted bearings jointly with
quadrotor dynamics. LOS velocity tracking, bearing-dependent speed demand,
and visual costs and constraints coordinate approach and visibility.
Compensated bearings and predicted regions of interest (ROIs) support detection.

The main contributions are as follows:
\begin{itemize}[leftmargin=*]
    \item An interception-oriented spherical-bearing NMPC that jointly
    optimizes desired closing speed and vehicle motion. LOS-aligned relative
    velocity objectives, visual costs, and soft FOV constraints coordinate
    rapid closing and target visibility under quadrotor dynamics and actuator
    limits.
    \item A spherical-bearing estimator with attitude and delay compensation
    that supplies current-time bearing feedback and absolute-time ROI
    predictions to support onboard operation during large attitude changes
    and target-scale variation.
    \item Paired controller comparisons over 30 simulation trials and outdoor
    flights involving long-range acquisition of a quasi-static small target
    and pursuit of a maneuvering target, evaluating controller performance
    and the operation of the integrated perception--control system.
\end{itemize}

Section~\ref{sec:system_problem} defines the system and interception problem.
Sections~\ref{sec:perception} and~\ref{sec:isbnmpc} present the perception front
end and ISB-NMPC, respectively. Section~\ref{sec:validation} reports simulation
and flight validation, followed by the discussion and conclusion.

\begin{figure}[!b]
    \centering
    \includegraphics[width=\columnwidth]{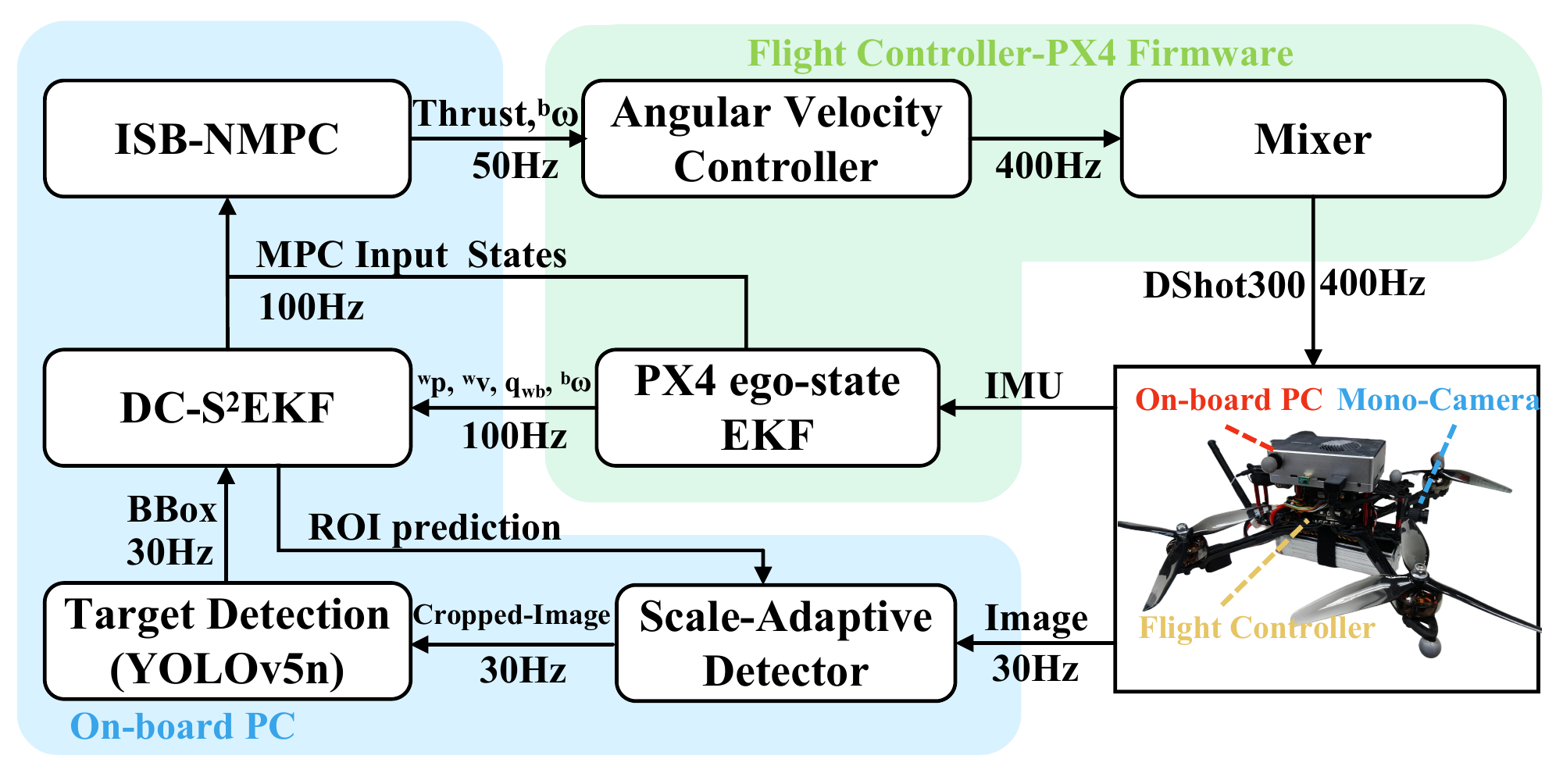}
    \caption{Onboard architecture integrating scale-adaptive detection,
    DC-$\mathbb S^2$EKF bearing estimation, ISB-NMPC, and PX4 actuation.}
    \label{fig:system_architecture}
    \par\smallskip
    {\footnotesize\raggedright\textsuperscript{1}\hspace{0.35em}YOLOv5 source code:
    \url{https://github.com/ultralytics/yolov5}.\par}
\end{figure}

\section{System Architecture and Problem Formulation}
\label{sec:system_problem}

\subsection{System Architecture}

The onboard system comprises a PX4 flight controller whose
extended Kalman filter (EKF) supplies ego-state, a strapdown
monocular USB
camera, and an NVIDIA Jetson Orin NX companion computer. As shown in
Fig.~\ref{fig:system_architecture}, $30\,\mathrm{Hz}$ images feed a
scale-adaptive YOLOv5n detector,\textsuperscript{1} and a
delay-compensated
$\mathbb S^2$ EKF (DC-$\mathbb S^2$EKF) fuses the detections with this
ego-state to provide current-time bearings and ROI
predictions, where
$\mathbb S^2:=\{\mathbf n\in\mathbb R^3:\|\mathbf n\|_2=1\}$ denotes the unit
two-sphere. ISB-NMPC runs at $50\,\mathrm{Hz}$ and sends collective-thrust and
body-rate commands to PX4.

\subsection{Coordinate Frames and Quadrotor Dynamics}
\label{sec:frames_dynamics}

\begin{figure}[!tbp]
\centering
\includegraphics[width=\columnwidth]{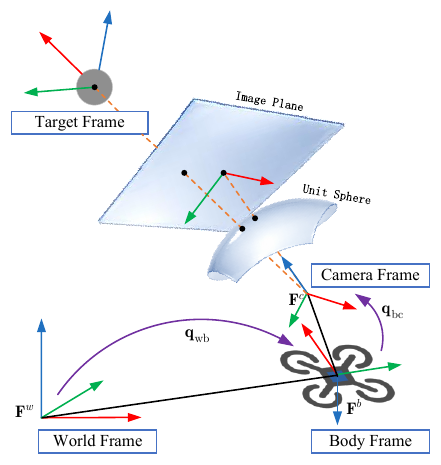}
\caption{
World, body, camera, and target frames, with spherical bearing
$\mathbf n_t\in\mathbb S^2$ and image-plane projection.
Axes $x$, $y$, and $z$ are red, green, and blue, respectively.
}
\label{fig:coordinate_frames}
\end{figure}

Fig.~\ref{fig:coordinate_frames} defines the world, body, camera, and target
frames $\mathcal F^w$, $\mathcal F^b$, $\mathcal F^c$, and $\mathcal F^t$.
The vector ${}^{a}\mathbf p_b$ points from the origin of $\mathcal F^a$ to
that of $\mathcal F^b$ and is expressed in $\mathcal F^a$; the unit quaternion
$\mathbf q_{ab}\in\mathbb S^3$ maps vectors from $\mathcal F^b$ to
$\mathcal F^a$ through $R(\mathbf q_{ab})\in SO(3)$. Quaternions are ordered
scalar-first, composition is denoted by $\otimes$, and
$\mathbf e_3=[0,0,1]^\top$. The fixed camera extrinsic $\mathbf q_{bc}$ and
vehicle attitude $\mathbf q_{wb}$ give
$\mathbf q_{wc}=\mathbf q_{wb}\otimes\mathbf q_{bc}$ and
$\mathbf q_{cw}=\mathbf q_{wc}^{-1}$.
Let ${}^{b}\mathbf p_c$ denote the camera position in the body frame.

Given the UAV and target positions ${}^{w}\mathbf p_b$ and ${}^{w}\mathbf p_t$, the
target position and unit bearing in the camera frame are
\begin{equation}
\begin{aligned}
    {}^{c}\mathbf p_t
    &=R(\mathbf q_{bc})^\top
      \left[
      R(\mathbf q_{wb})^\top
      \left({}^{w}\mathbf p_t-{}^{w}\mathbf p_b\right)
      -{}^{b}\mathbf p_c
      \right],\\
    \mathbf n_t
    &=\frac{{}^{c}\mathbf p_t}{\|{}^{c}\mathbf p_t\|}
      \in\mathbb S^2.
\end{aligned}
    \label{eq:target_camera_geometry}
\end{equation}

\stepcounter{figure}
\begin{figure*}[!t]
    \centering
    \includegraphics[width=\textwidth]{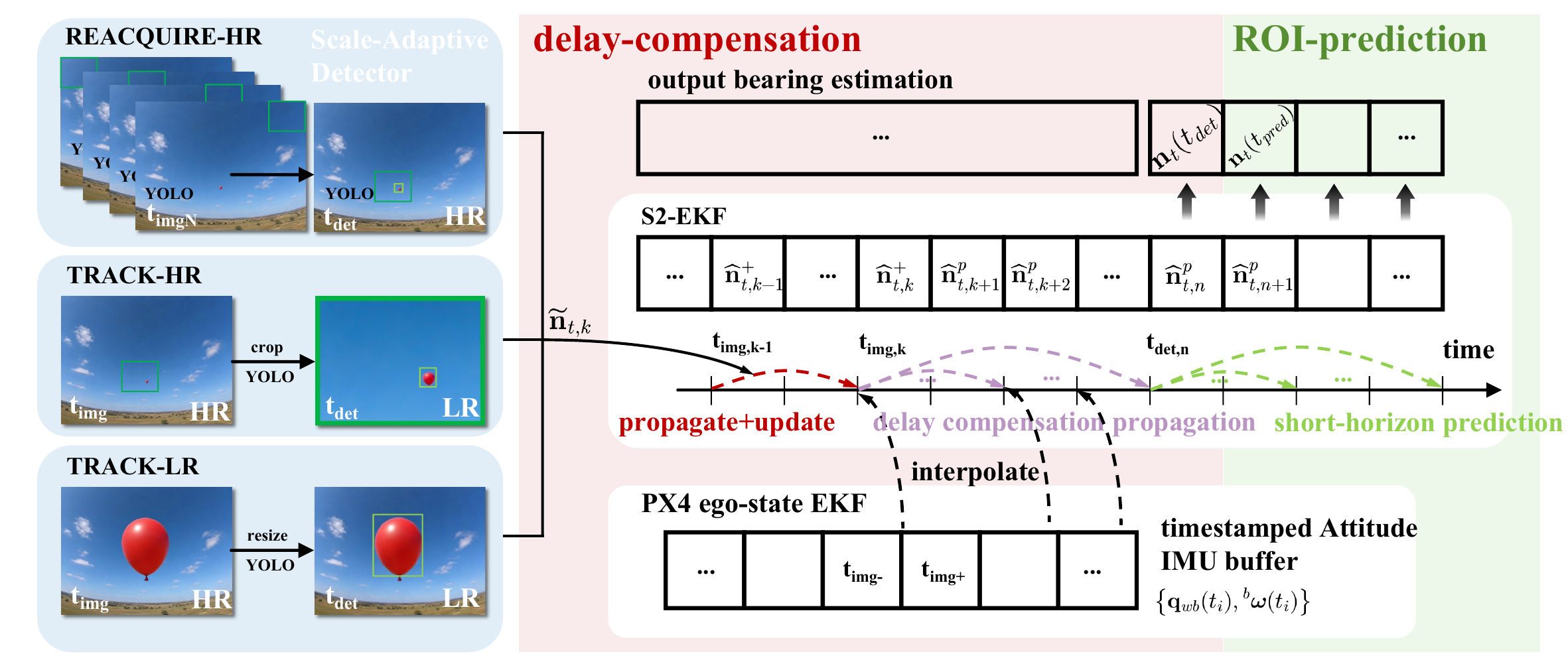}
    \caption{Detection modes and timing of delay compensation
    and ROI prediction. Superscript $p$ denotes forward predictions from the
    latest posterior, without modifying the EKF state.}
    \label{fig:perception_timing}
\end{figure*}

The controller uses the mass-normalized collective thrust $T$ and body angular
velocity ${}^{b}\boldsymbol\omega$ as inputs. The quadrotor model is
\begin{equation}
    {}^{w}\dot{\mathbf p}_b={}^w\mathbf v_b,
    \qquad
    {}^{w}\dot{\mathbf v}_b
    =T R(\mathbf q_{wb})\mathbf e_3-g\mathbf e_3,
    \label{eq:quad_trans_dynamics}
\end{equation}
with attitude kinematics
\begin{equation}
    \dot{\mathbf q}_{wb}
    =\frac{1}{2}\mathbf q_{wb}\otimes
    \begin{bmatrix}0, & {}^{b}\boldsymbol\omega^\top\end{bmatrix}^\top.
    \label{eq:attitude_dynamics}
\end{equation}
The camera is rigidly attached to the body. Because the PX4 body-rate loop is
substantially faster than the NMPC update, the commanded body rate is treated
as a directly tracked input; rotational rigid-body, motor, and propeller
dynamics are not included in the prediction model.

\subsection{Visual Interception Problem}
\label{sec:visual_interception_problem}

We consider a quadrotor intercepting a non-cooperative moving target using a
rigidly mounted monocular camera, as illustrated in
Fig.~\ref{fig:interception_process}.
The target does not communicate its state to the
interceptor. The onboard front end always supplies a compensated bearing; when
an onboard target-state estimate is available, its position and velocity are
also used by the controller. In bearing-only operation, the controller
initializes the prediction with a nominal range and zero target velocity.
Let
\begin{equation}
    {}^{w}\mathbf r_t
    ={}^{w}\mathbf p_t-{}^{w}\mathbf p_b,
    \qquad
    r=\|{}^{w}\mathbf r_t\|,
    \qquad
    {}^{w}\mathbf n_t
    =\frac{{}^{w}\mathbf r_t}{r},
    \label{eq:relative_interception_geometry}
\end{equation}
denote the relative displacement, range, and line-of-sight (LOS) direction
from the quadrotor to the target. With
${}^{w}\mathbf v_r={}^w\mathbf v_b-{}^w\mathbf v_t$, the physical
instantaneous closing speed is
\begin{equation}
    s_c=-\dot r
    ={}^{w}\mathbf n_t^\top{}^{w}\mathbf v_r,
    \label{eq:physical_closing_speed}
\end{equation}
where $s_c>0$ indicates that the range is decreasing.

Visibility is expressed using the camera-frame bearing
$\mathbf n_t=[n_x,n_y,n_z]^\top$ defined in
\eqref{eq:target_camera_geometry}. Its normalized image coordinate is
$\boldsymbol\pi(\mathbf n_t)=[n_x/n_z,\ n_y/n_z]^\top$ for $n_z>0$.
The ideal visibility requirement is
$\boldsymbol\pi_{\min}\leq\boldsymbol\pi(\mathbf n_t)
\leq\boldsymbol\pi_{\max}$, where the bounds are determined by the usable
camera field of view.

LOS-constrained interceptability couples target-directed motion with
visual contact~\cite{yangLineofSightConstrainedMulticopterInterceptability2025a,zhangInterceptabilityAnalysisLineofsight2026}.
At fixed geometry and target velocity, Cauchy--Schwarz gives
\begin{equation}
\begin{aligned}
 s_c&={}^w\mathbf n_t^\top{}^w\mathbf v_b
       -{}^w\mathbf n_t^\top{}^w\mathbf v_t\\
    &\leq s-{}^w\mathbf n_t^\top{}^w\mathbf v_t,
       \qquad \|{}^w\mathbf v_b\|_2=s.
\end{aligned}
\label{eq:instantaneous_closing_bound}
\end{equation}

Equality holds for ${}^w\mathbf v_b=s\,{}^w\mathbf n_t$: LOS alignment
maximizes instantaneous closing, not minimum-time capture.
With estimated target velocity, the nominal relative reference is
${}^w\mathbf v_{r,d}=s\,{}^w\mathbf n_t$, giving
${}^w\mathbf v_{b,d}=\hat{\mathbf v}_t+s\,{}^w\mathbf n_t$;
bearing-only operation sets $\hat{\mathbf v}_t=\mathbf0$.

\textbf{Dynamic coupling.} Differentiating
\eqref{eq:physical_closing_speed} using
\eqref{eq:quad_trans_dynamics}, for differentiable target velocity and $r>0$,
gives
\begin{equation}
\begin{aligned}
\dot s_c
 &= {}^w\mathbf n_t^\top
    (T\mathbf b_3-g\mathbf e_3-{}^w\dot{\mathbf v}_t)
    -\frac{\|\mathbf v_\perp\|_2^2}{r},\\
\mathbf b_3&=R(\mathbf q_{wb})\mathbf e_3,\qquad
\mathbf v_\perp=(\mathbf I_3-{}^w\mathbf n_t{}^w\mathbf n_t^\top)
{}^w\mathbf v_r.
\end{aligned}
\label{eq:closing_acceleration_coupling}
\end{equation}
The negative transverse-motion term follows from
${}^w\dot{\mathbf n}_t=-\mathbf v_\perp/r$.
Increasing thrust improves closing acceleration only when
${}^w\mathbf n_t^\top\mathbf b_3>0$; changing this alignment rotates the
strapdown camera. The same body rate enters the rotational term of
\eqref{eq:bearing_kinematics_s2}. Thus thrust and attitude cannot independently
set closing and visibility. This motivates predicting both and optimizing
the speed demand rather than imposing a fixed maximal speed.

The soft relaxation used by the controller is introduced
in Sec.~\ref{sec:nmpc_objective_constraints}. Meanwhile, the control input
must remain in the feasible set
\begin{equation}
    \mathcal U=
    \left\{(T,{}^{b}\boldsymbol\omega)\ \middle|
    T_{\min}\leq T\leq T_{\max},\;
    |\omega_i|\leq\bar\omega_i,\ i\in\{x,y,z\}\right\}.
    \label{eq:feasible_input_set}
\end{equation}

\textbf{Interception problem.} Given bearing feedback and available
target-state estimates, choose $(T,{}^b\boldsymbol\omega)\in\mathcal U$
to reach $r(t_f)\leq r_{\mathrm{int}}$ rapidly while retaining the target
within the FOV, without a preplanned target trajectory.
Performance is assessed jointly by success rate, interception time among
successful trials, and target-visible time: fast closing alone is insufficient.
Section~\ref{sec:nmpc_objective_constraints} constructs a finite-horizon
surrogate, not a guaranteed minimum-time interception solution.
In particular, the non-cooperative target's future motion is unknown a priori.

\section{Onboard Visual Perception Front End}
\label{sec:perception}

The visual front end converts timestamped images and ego-motion into
native-image bounding boxes, current-time unit bearings, residual bearing
motion, and future ROI predictions for detection and control.

\subsection{Bearing Estimation, Temporal Filtering, and Delay Compensation}

The estimator preserves unit bearings during camera rotation and uses
two-dimensional tangent innovations (Fig.~\ref{fig:dcs2_ekf_geometry}).
Superscripts $r$, $-$, and $+$ denote rotation-transported, prior, and
posterior quantities; $k$ indexes filter samples, a hat denotes an estimate,
and $t$ identifies the target.

\begin{figure}[!htbp]
    \centering
    \includegraphics[width=\columnwidth]{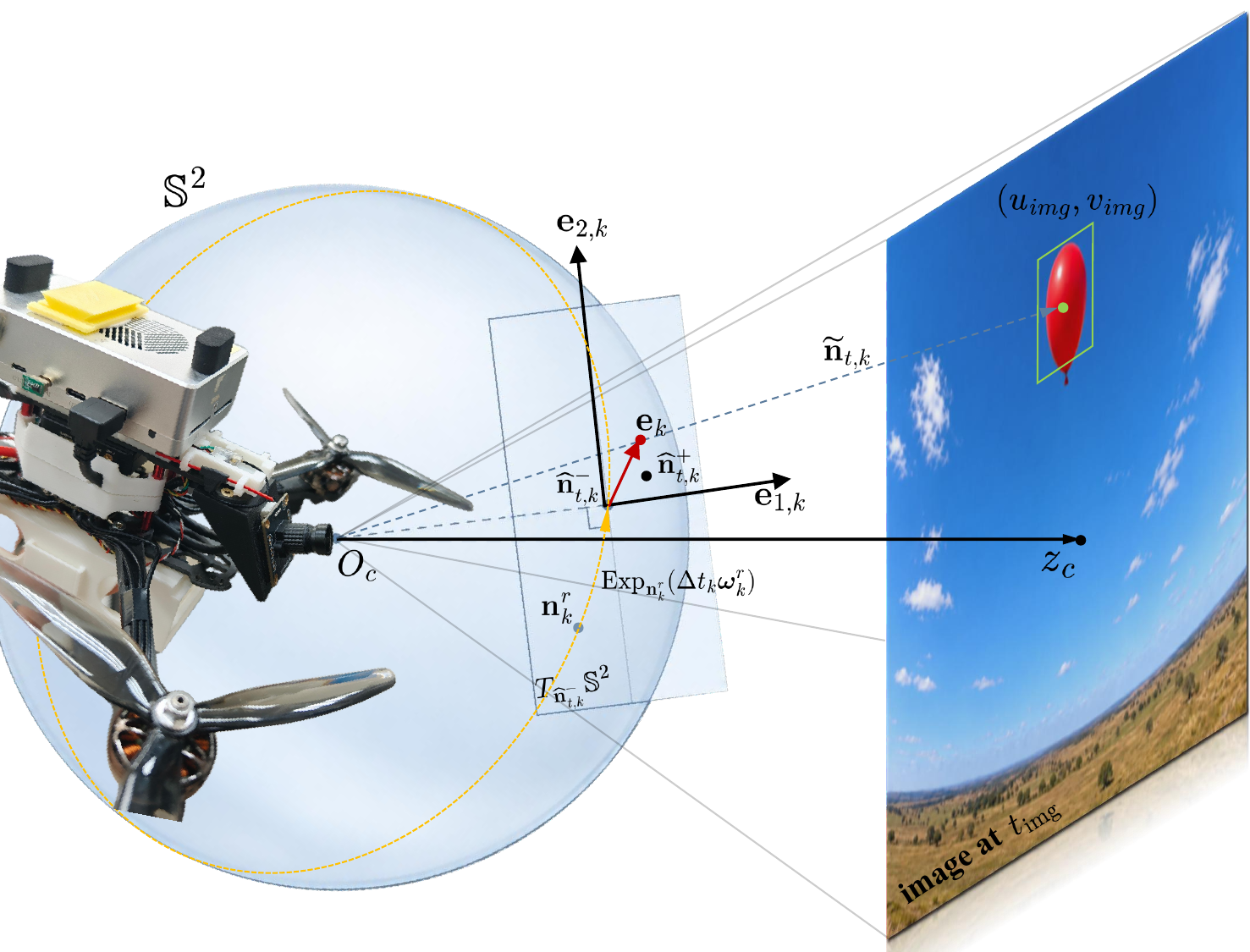}
    \addtocounter{figure}{-2}
    \caption{DC-$\mathbb S^2$EKF geometry: spherical back-projection,
    rotation compensation, propagation, and tangent-space correction.
    Bearings and tangent bases are shown schematically in the camera frame
    at measurement time.}
    \label{fig:dcs2_ekf_geometry}
    \stepcounter{figure}
\end{figure}

At image time $t_{\mathrm{img}}$, the bounding-box center
$(u_{\mathrm{img}},v_{\mathrm{img}})$ and camera intrinsics $\mathbf K$
give the unit bearing

\begin{equation}
    \widetilde{\mathbf n}_{t}(t_{\mathrm{img}})
    =\frac{\mathbf K^{-1}
    [u_{\mathrm{img}},v_{\mathrm{img}},1]^\top}
    {\|\mathbf K^{-1}
    [u_{\mathrm{img}},v_{\mathrm{img}},1]^\top\|}
    \in\mathbb S^2.
    \label{eq:measured_spherical_bearing}
\end{equation}

The DC-$\mathbb S^2$EKF separates camera rotation from residual LOS motion
using the bearing $\widehat{\mathbf n}_t$ and tangent velocity
$\widehat{\boldsymbol\omega}_n\in T_{\widehat{\mathbf n}_t}\mathbb S^2$.
Local tangent coordinates follow the manifold-estimation
principle in~\cite{hertzbergManifoldFusion2013}.
For two times $t_a$ and $t_b$, define the camera-frame relative rotation
$\mathbf R_{c(t_b)c(t_a)}=R(\mathbf q_{wc}(t_b))^\top
R(\mathbf q_{wc}(t_a))$.
For estimator times $t_k$ and $\Delta t_k=t_k-t_{k-1}$, let
$\mathbf R_k=\mathbf R_{c(t_k)c(t_{k-1})}$. Propagation compensates camera
rotation before integrating residual motion:

\begin{equation}
\begin{aligned}
    \mathbf n_k^r&=\mathbf R_k\widehat{\mathbf n}_{t,k-1},
    &\boldsymbol\omega_k^r
        &=\Pi_{\mathbf n_k^r}\mathbf R_k
          \widehat{\boldsymbol\omega}_{n,k-1},\\
    \widehat{\mathbf n}_{t,k}^{-}
        &=\operatorname{Exp}_{\mathbf n_k^r}
          (\Delta t_k\boldsymbol\omega_k^r),
    &\Pi_{\mathbf n}&=\mathbf I_3-\mathbf n\mathbf n^\top.
\end{aligned}
    \label{eq:s2_filter_propagation}
\end{equation}
The predicted residual velocity is projected into the new tangent plane,
$\widehat{\boldsymbol\omega}_{n,k}^{-}
=\Pi_{\widehat{\mathbf n}_{t,k}^{-}}\boldsymbol\omega_k^r$.
This residual tangent velocity differs from the total spin-free angular
velocity $\mathbf n_t\times\dot{\mathbf n}_t$ used in the control cost.

With tangent basis $\mathbf B_k=[\mathbf e_{1,k},\mathbf e_{2,k}]$, the local
error comprises two bearing and two residual-velocity coordinates. A
constant-velocity model propagates its covariance using $\Delta t_k$,
residual-acceleration variance $\sigma_a^2$, and IMU rotation uncertainty
$\sigma_g^2$. The update is
\begin{equation}
\begin{aligned}
 \mathbf y_k
 &=\mathbf B_k^\top
   \operatorname{Log}_{\widehat{\mathbf n}_{t,k}^{-}}
   (\widetilde{\mathbf n}_{t,k}),\\
 \mathbf K_k
 &=\mathbf P_k^-\mathbf H^\top
   (\mathbf H\mathbf P_k^-\mathbf H^\top+\mathbf R_k^n)^{-1},\\
 \begin{bmatrix}\delta\boldsymbol\theta_k\\
                 \delta\boldsymbol\omega_k\end{bmatrix}
 &=\mathbf K_k\mathbf y_k,\\
 \widehat{\mathbf n}_{t,k}^{+}
 &=\operatorname{Exp}_{\widehat{\mathbf n}_{t,k}^{-}}
   (\mathbf B_k\delta\boldsymbol\theta_k),\\
 \widehat{\boldsymbol\omega}_{n,k}^{+}
 &=\Pi_{\widehat{\mathbf n}_{t,k}^{+}}
   (\widehat{\boldsymbol\omega}_{n,k}^{-}
    +\mathbf B_k\delta\boldsymbol\omega_k),
 \qquad \mathbf H=[\mathbf I_2\ \mathbf0].
\end{aligned}
\label{eq:s2_ekf_compact_update}
\end{equation}
The angular covariance $\mathbf R_k^n$ follows from pixel covariance and the
back-projection Jacobian, with a noise floor set by confidence, box area,
boundary proximity, and temporal consistency through
$\rho_k\in[\rho_{\min},1]$. Joseph-form covariance update and basis transport
maintain consistency; the exponential map preserves bearing norm.

For a detection received at $t_{\mathrm{det}}$, the measurement update is
applied at $t_{\mathrm{img}}$ and propagated to the current time using
buffered attitude and IMU data, with the relative rotation
$\mathbf R_{c(t_{\mathrm{det}})c(t_{\mathrm{img}})}$ defined above.
Propagation to controller queries supplies current-time bearings and to image
acquisition times supplies ROI predictions. Missing timestamps use a calibrated
delay; stale IMU data or prolonged detection loss triggers reacquisition.

\subsection{Multi-Resolution Detection and ROI-Based Tracking}

A TensorRT-accelerated YOLO backend switches among tiled high-resolution
reacquisition, prediction-centered high-resolution tracking, and downsampled
full-frame tracking using hysteretic size and confidence thresholds. Valid
predictions bridge brief dropouts; repeated loss triggers reacquisition.
Figure~\ref{fig:perception_timing} details these modes (left) and the
timestamp-based delay compensation and ROI prediction process (right).

\begingroup
\setlength{\abovedisplayskip}{3pt plus 1pt minus 1pt}
\setlength{\belowdisplayskip}{3pt plus 1pt minus 1pt}
\setlength{\belowdisplayshortskip}{3pt plus 1pt minus 1pt}
\section{Interception-Oriented Spherical-Bearing NMPC}
\label{sec:isbnmpc}

The controller receives the current ego-state, compensated camera-frame
bearing, and the available target-motion estimate. At every update it builds a
short-horizon prediction, schedules the closing-speed reference from the
current LOS error, solves the constrained optimal-control problem, and applies
only the first thrust and body-rate command.

\subsection{Spherical-Bearing Model and Prediction Dynamics}

The rotation-equivariant unit bearing separates direction from
range~\cite{hamelVisualServoingUnderactuated2002}. Its $SO(3)$ lift enables
joint visual and attitude prediction~\cite{qinPerceptionawareImagebasedVisual2023}.

\subsubsection{SO(3) Lifting of the Filtered Bearing}

The front end supplies $\mathbf n_t\in\mathbb S^2$ through the measurement,
filtering, and delay-compensation process of Sec.~\ref{sec:perception}.
For forward bearings $n_z=\mathbf e_3^\top\mathbf n_t>0$, the minimal rotation
from $\mathbf e_3$ to $\mathbf n_t$ has the nonsingular unit quaternion
\begin{equation}
{}^{c}\mathbf f
=\frac{[1+n_z,(\mathbf e_3\times\mathbf n_t)^\top]^\top}
{\sqrt{2(1+n_z)}}\in\mathbb S^3.
\end{equation}
It recovers the bearing as $\mathbf n_t=R({}^{c}\mathbf f)\mathbf e_3$;
the camera-frame target position is ${}^{c}\mathbf p_t=r\mathbf n_t$,
with $r=\|{}^{c}\mathbf p_t\|_2$.

\subsubsection{Spherical-Bearing Kinematics}

To propagate ${}^{c}\mathbf f$, account for relative target translation and
camera rotation. Let ${}^{c}\boldsymbol\omega_{wc}$ denote the angular
velocity of the camera frame with respect to the world frame, expressed in
$\mathcal F^c$. Differentiating the normalized bearing gives
\begin{equation}
    \dot{\mathbf n}_t
    =
    -
    \left[
    {}^{c}\boldsymbol\omega_{wc}
    \right]_{\times}
    \mathbf n_t
    +
    \frac{1}{r}
    \left(
    \mathbf I_3
    -
    \mathbf n_t \mathbf n_t^{\top}
    \right)
    {}^{c}\mathbf v_{t/b}.
    \label{eq:bearing_kinematics_s2}
\end{equation}
Only transverse relative velocity changes bearing. Since
$\dot{\mathbf n}_t\in T_{\mathbf n_t}\mathbb S^2$, define the spin-free angular
velocity
\begin{equation}
    {}^{c}\boldsymbol\omega_{n}
    =
    \mathbf n_t
    \times
    \dot{\mathbf n}_t,
    \qquad
    \dot{\mathbf n}_t
    =
    \left[
    {}^{c}\boldsymbol\omega_{n}
    \right]_{\times}
    \mathbf n_t.
\label{eq:bearing_angular_velocity}
\end{equation}
With $\mathbf n_t=R({}^{c}\mathbf f)\mathbf e_3$, the quaternion kinematics
used in the prediction model are
\begin{equation}
    {}^{c}\dot{\mathbf f}
    =
    \frac{1}{2}
    \begin{bmatrix}
        0\\
        {}^{c}\boldsymbol\omega_n
    \end{bmatrix}
    \otimes
    {}^{c}\mathbf f.
    \label{eq:quaternion_bearing_kinematics}
\end{equation}
This directly propagates the SO(3)-lifted visual feature while preserving
the unit-norm constraint of the quaternion representation.

\subsubsection{NMPC State and Control Input}

The prediction state combines vehicle motion, the visual feature, and
interception variables:
\begin{equation}
\mathbf x
=[{}^{w}\mathbf p_b^\top,\mathbf q_{wb}^\top,{}^{w}\mathbf v_b^\top,
{}^{c}\mathbf f^\top,r,s,z_u,z_v]^\top\in\mathbb R^{18}.
\end{equation}
Vehicle variables follow Sec.~\ref{sec:frames_dynamics}, with the current
position as the local prediction origin. The desired closing-speed state $s$
is distinct from physical closing speed $-\dot r$. Auxiliary accumulators
$z_u,z_v$ retain the FOV slacks as decision inputs in the generated Automatic
Control and Dynamic Optimization (ACADO)~\cite{houskaACADO2011} solver and are
not physical states.

The control input is defined as
\begin{equation}
\mathbf u
=[T,{}^b\omega_x,{}^b\omega_y,{}^b\omega_z,
u_s,\epsilon_u,\epsilon_v]^\top\in\mathbb R^7.
\end{equation}
Here $T$ and ${}^{b}\boldsymbol\omega$ are the thrust and body-rate inputs
defined in Sec.~\ref{sec:frames_dynamics}; $u_s$ drives desired speed, and
$\epsilon_u,\epsilon_v$ relax horizontal and vertical FOV bounds, with
$\dot z_u=\epsilon_u$ and $\dot z_v=\epsilon_v$.

\subsubsection{Prediction Dynamics}

The calibrated camera rotation gives the angular and relative velocities
used in \eqref{eq:bearing_kinematics_s2}--\eqref{eq:quaternion_bearing_kinematics}:
\begin{align}
{}^{c}\boldsymbol\omega_{wc}
 &=R(\mathbf q_{bc})^\top{}^{b}\boldsymbol\omega,\\
{}^{w}\mathbf v_r
 &={}^{w}\mathbf v_b-{}^{w}\hat{\mathbf v}_t,
 \label{eq:relative_velocity_world}\\
{}^{c}\mathbf v_{t/b}
 &=-R(\mathbf q_{cw}){}^{w}\mathbf v_r.
 \label{eq:relative_velocity_camera}
\end{align}
These differential kinematics neglect the small camera--body translational
lever arm.

The range state obeys $\dot r=-s_c$ from
\eqref{eq:physical_closing_speed}. The desired closing-speed state is driven
by the optimization input as $\dot s=u_s$.
At prediction node $i$, the world-frame bearing is
${}^{w}\mathbf n_{t,i}=R(\mathbf q_{wb,i})R(\mathbf q_{bc})R({}^{c}\mathbf f_i)\mathbf e_3$.
It evolves with the predicted attitude and visual feature; only the reference
${}^{w}\mathbf n_{t,0}$ in the objective is frozen. The velocity-tracking cost
below encourages the physical closing speed to approach $s$.

Together, the vehicle, attitude, bearing, range, and desired-speed dynamics
define the continuous-time prediction model.

\subsection{NMPC Objective and Constraints}
\label{sec:nmpc_objective_constraints}

\subsubsection{Objective}

Equation~\eqref{eq:closing_acceleration_coupling} motivates the
objective: LOS velocity tracking encourages approach and discourages off-axis
motion, while visual costs account for camera rotation during acceleration.
Optimizing $s$ permits adjustment of the approach demand under vehicle
constraints. Each solve holds the estimated target velocity, measured LOS
${}^w\mathbf n_{t,0}$, and bearing-based reference $s_d$ constant.
The interception residual promotes LOS-aligned closing
motion~\cite{yangHighSpeedInterceptionMulticopter2025a} through
$\mathbf e_{\mathrm{int}}={}^w\mathbf v_r-s\,{}^w\mathbf n_{t,0}$.
To moderate the speed demand near the FOV boundary, $s_d$ is scheduled
using the horizontal and vertical LOS errors. For
$\mathbf n_t=[n_x,n_y,n_z]^\top$, the implementation uses
\begin{equation}
    \begin{aligned}
        e_h&=\frac{n_x}{\sqrt{n_x^2+n_z^2+\varepsilon^2}},\\
        e_v&=\frac{n_y}{\sqrt{n_y^2+n_z^2+\varepsilon^2}},\\
        \alpha_{\mathrm{los}}
        &=
        \exp[-(k_h e_h^2+k_v e_v^2)],\\
        s_d
        &=
        s_{\min}
        +
        \alpha_{\mathrm{los}}(s_{\max}-s_{\min}),
    \end{aligned}
    \label{eq:los_scheduled_closing_speed}
\end{equation}
where $k_h,k_v\geq0$ determine the horizontal and vertical scheduling
sensitivity. Using the current bearing prevents predicted errors from
reducing the optimizer's own speed reference.
The running interception cost is
\begin{equation}
    \mathcal L_{\mathrm{int}}
    =
    \left\|\mathbf e_{\mathrm{int}}\right\|_{\mathbf Q_{\mathrm{int}}}^{2}
    +
    q_s(s-s_d)^2.
    \label{eq:interception_cost}
\end{equation}

Tracking $s_d$ encourages rapid approach near the optical axis;
optimizing $s$ allows compromise with visual and actuation requirements.

Consistent with the field-of-view constraint, we formulate the visual objective
directly in the normalized image plane. Let
$\boldsymbol{\pi}_\varepsilon(\mathbf n_t)
=[n_x/(n_z+\varepsilon),n_y/(n_z+\varepsilon)]^\top$ denote the regularized normalized
image projection.
To measure the distance from the target projection to the image center, we
define
\begin{equation}
\begin{aligned}
\boldsymbol{\pi}_c
&=\frac{\boldsymbol{\pi}_{\max}+\boldsymbol{\pi}_{\min}}{2},
&\boldsymbol{\rho}
&=\frac{\boldsymbol{\pi}_{\max}-\boldsymbol{\pi}_{\min}}{2},\\
\mathbf e_{\mathrm{vis}}
&=\operatorname{diag}(\boldsymbol{\rho})^{-1}
\left(\boldsymbol{\pi}_\varepsilon(\mathbf n_t)-\boldsymbol{\pi}_c\right).
\end{aligned}
\label{eq:visual_residual}
\end{equation}

Since velocity tracking alone cannot prevent camera-induced target
displacement, the visual cost penalizes image offset and bearing motion:
\begin{equation}
\mathcal L_{\mathrm{vis}}
=\left\|\mathbf e_{\mathrm{vis}}\right\|_{\mathbf Q_{\mathrm{vis}}}^{2}
+\left\|{}^{c}\boldsymbol\omega_n\right\|_{\mathbf Q_{\mathrm{jit}}}^{2},
\label{eq:visual_cost}
\end{equation}
where ${}^{c}\boldsymbol\omega_n$ is defined in
\eqref{eq:bearing_angular_velocity}.

The regularization term
$\mathcal L_{\mathrm{reg}}=
\|\mathbf u-\mathbf u_{\mathrm{ref}}\|_{\mathbf R_u}^{2}$ penalizes excessive
actuation and aggressive closing-speed changes,
where $\mathbf u_{\mathrm{ref}}=[g,0,0,0,0,0,0]^\top$ is the nominal hover
input. In particular, this term penalizes thrust deviation from gravity, body
rates, closing-speed rate, and both FOV slacks. The terminal cost and its
cross-track residual are
\begin{equation}
\begin{aligned}
    \mathcal L_N
    &=
    \left\|\mathbf e_{\mathrm{vis},N}\right\|_{\mathbf Q_{\mathrm{vis},N}}^{2}
    +
    q_{\ell,N}e_{\ell,N}^2,\\
    e_{\ell,N}
    &=
    \left\|
    \left(
    {}^{w}\mathbf p_N
    -
    {}^{w}\mathbf p_{\ell,0}
    \right)
    \times
    {}^{w}\mathbf n_{t,0}
    \right\|_2.
\end{aligned}
    \label{eq:terminal_cost}
\end{equation}
The residual $e_{\ell,N}$ penalizes the perpendicular deviation of the predicted terminal position from
the line passing through the current local prediction origin
${}^{w}\mathbf p_{\ell,0}$ along the frozen LOS direction. This soft penalty
discourages off-axis motion while the underactuated vehicle accelerates
toward the desired closing speed.

\subsubsection{Constraints}

The input set and visibility requirements of
Sec.~\ref{sec:visual_interception_problem} become actuator bounds and soft FOV
constraints, supplemented by range and speed bounds:
\begin{equation}
\begin{aligned}
    \mathbf u_{\min}
    &\leq
    \mathbf u
    \leq
    \mathbf u_{\max},\\
    0
    &\leq
    s
    \leq
    v_{\ell,\max},\\
    r
    &\geq
    r_{\min},\qquad n_z\geq n_{z,\min},\\
    \boldsymbol\pi_{\min}-\boldsymbol\epsilon
    &\leq
    \boldsymbol\pi_\varepsilon(\mathbf n_t)
    \leq
    \boldsymbol\pi_{\max}+\boldsymbol\epsilon,\\
    &\mathbf0\leq\boldsymbol\epsilon=[\epsilon_u,\epsilon_v]^\top
    \leq\boldsymbol\epsilon_{\max}.
\end{aligned}
\label{eq:system_constraints}
\end{equation}
Here $\boldsymbol\pi_{\min}$ and $\boldsymbol\pi_{\max}$ denote the lower and
upper normalized image bounds induced by the camera field of view. The bounded,
nonnegative slack vector $\boldsymbol\epsilon$ makes the FOV constraint soft,
and its large running penalty discourages violations. The additional lower
bound on $n_z$ keeps the predicted target in front of the camera. Since
$\dot s=u_s$, the corresponding entries of $\mathbf u_{\min}$ and
$\mathbf u_{\max}$ impose $|u_s|\leq a_{\ell,\max}$.

\subsubsection{Discrete Optimal Control Problem}

Let $\Delta t$ denote the prediction discretization interval, and let
$\mathbf F(\cdot)$ be the discretized continuous-time prediction model. Define
$\mathbf X=\{\mathbf x_i\}_{i=0}^{N}$ and
$\mathbf U=\{\mathbf u_i\}_{i=0}^{N-1}$. At each control update, the following
finite-horizon problem is solved:

\begin{equation}
\label{eq:discrete_nmpc_problem}
\begin{aligned}
    \min_{\mathbf X,\mathbf U}
    \quad &
    \sum_{i=0}^{N-1}
    \left(
    \mathcal L_{\mathrm{int},i}
    +
    \mathcal L_{\mathrm{vis},i}
    +
    \mathcal L_{\mathrm{reg},i}
    \right)
    +
    \mathcal L_N
    \\
    \mathrm{s.t.}
    \quad &
    \mathbf x_0=\hat{\mathbf x},
    \\
    &
    \mathbf x_{i+1}
    =
    \mathbf F(\mathbf x_i,\mathbf u_i,\hat{\mathbf v}_{t}),
    \\
    &
    \mathbf h_c(\mathbf x_i,\mathbf u_i)\leq \mathbf 0,
    \qquad
    i=0,\ldots,N-1.
\end{aligned}
\end{equation}

\textbf{Existence and scope of optimality.}
For fixed state and target-velocity estimates, assume a nonempty bounded input
set, a projection denominator separated from zero, and at least one feasible
control sequence. Continuity of $\mathbf F$ and the costs, together with the
closed constraints, makes the feasible image of the compact input set compact;
the Weierstrass theorem therefore guarantees a global minimizer of
\eqref{eq:discrete_nmpc_problem}. This is an existence result only: the
nonconvex dynamics, projection, and constraints mean that the real-time solver
generally returns a local solution. Bounded FOV slack improves practical
feasibility but does not guarantee recursive feasibility, which would require
a compatible invariant terminal set and controller.

\subsection{Real-Time Implementation}

ISB-NMPC runs at 50~Hz in Robot Operating System (ROS) using an
ACADO-generated real-time-iteration solver
with qpOASES~\cite{ferreauQpOASES2014}. Ten $0.05$~s intervals form the
$0.5$~s horizon; Gauss--Newton approximation, implicit Gauss--Legendre
integration, full condensing, and warm starts are used. With seven inputs and
$N=10$, the condensed subproblem has at most 70 control increments and dense
linear algebra scales as $\mathcal O((7N)^3)$.
Typical onboard solve time on the Jetson Orin NX is about
10~ms versus the 20~ms control period, not a worst-case guarantee.
A failed solve briefly reuses the last valid command;
persistent failure resets the warm start and commands level hover.

\begin{figure*}[!t]
    \centering
    \includegraphics[width=\textwidth]{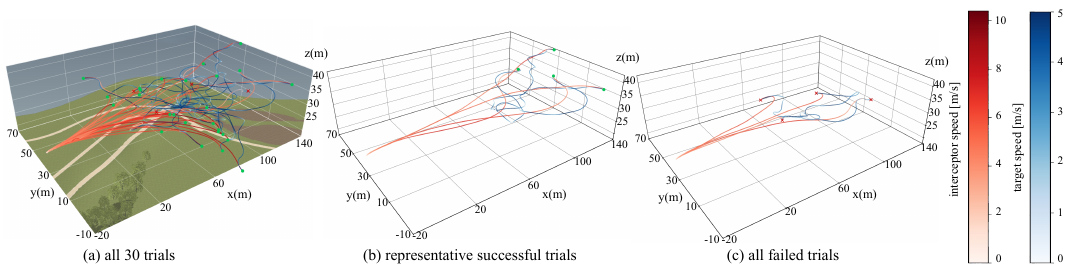}
    \caption{ISB-NMPC simulation trajectories:
    (a) all 30 trials, (b) selected successful trials, and (c) all failed trials.
    Colors indicate interceptor and target speeds; green markers denote
    interception and red crosses denote failure.}
    \label{fig:sim_controller_comparison}
\end{figure*}

\par\endgroup
\section{Simulation and Experimental Validation}
\label{sec:validation}

Simulation isolates controller performance with ground-truth target inputs;
outdoor flights evaluate the integrated bearing-only system.

\subsection{Simulation Validation and Comparison}

Thirty paired trials use ROS Noetic, Gazebo Classic 11, and PX4
software-in-the-loop at 50~Hz with identical platform settings and random
seeds. Trials start near 30~m altitude and 100~m separation. The 1~m-radius
target follows a three-dimensional random walk bounded by 5~m/s and
1.5~m/s$^2$, with direction updated every 3~s. At 30~Hz, ground-truth target
inputs isolate controller performance. Simulations run on an AMD Ryzen 9
9950X3D central processing unit and an NVIDIA RTX 5080 graphics processing
unit.

\begin{table}[!htbp]
    \centering
    \caption{Simulation comparison of interception controllers over 30 paired
    random-walk target-motion trials.}
    \label{tab:controller_comparison}
    \scriptsize
    \renewcommand{\arraystretch}{1.08}
    \setlength{\tabcolsep}{2.2pt}
    \resizebox{\columnwidth}{!}{%
    \begin{tabular}{>{\raggedright\arraybackslash}p{0.40\columnwidth}cccc}
        \toprule
        Metric & \makecell{ISB-\\NMPC} & PS-LOS & \makecell{LOS-\\IBVS}
        & \makecell{Back-\\stepping} \\
        \midrule
        \multicolumn{5}{l}{\emph{Interception performance}} \\
        Success rate (\%) $\uparrow$ & \makecell{\textbf{86.7}} & 63.3 & 6.7 & 33.3 \\
        Interception time (s) $\downarrow$ & $25.75\pm7.10$ & $35.40\pm15.28$ & \makecell{\textbf{$15.51\pm0.57$}} & $37.37\pm5.55$ \\
        Minimum target range (m) $\downarrow$ & \makecell{\textbf{$0.55\pm0.55$}} & $1.40\pm0.33$ & $13.26\pm11.67$ & $13.59\pm29.76$ \\
        Mean closing speed ($\mathrm{m\,s^{-1}}$) $\uparrow$ & $4.25\pm1.04$ & $3.02\pm1.21$ & \makecell{\textbf{$5.81\pm1.09$}} & $2.11\pm1.66$ \\
        \midrule
        \multicolumn{5}{l}{\emph{Target visibility}} \\
        Target-visible time (\%) $\uparrow$ & \makecell{\textbf{$99.11\pm1.21$}} & $56.10\pm9.97$ & $58.81\pm16.10$ & $63.87\pm14.84$ \\
        RMS normalized image error $\downarrow$ & \makecell{\textbf{$0.82\pm0.76$}} & $0.99\pm0.05$ & $1.24\pm0.41$ & $1.06\pm0.23$ \\
        \midrule
        \multicolumn{5}{l}{\emph{Control execution}} \\
        RMS body-rate variation ($\mathrm{rad\,s^{-2}}$) $\downarrow$
                                                            & $4.06\pm2.88$ & $9.58\pm2.48$ & \makecell{\textbf{$0.75\pm0.22$}} & $2.87\pm1.80$ \\
        \bottomrule
    \end{tabular}}
    \vspace{0.3ex}

    {\scriptsize Mean $\pm$
    standard deviation (SD) over all
    trials; interception time uses successful
    trials only. Image error is normalized by safe FOV radii.
   {$\uparrow$/$\downarrow$: higher/lower is preferred;
    closing speed must be assessed jointly with success and visibility.}}
\end{table}

\begin{figure}[!htbp]
    \centering
    \includegraphics[width=\columnwidth]{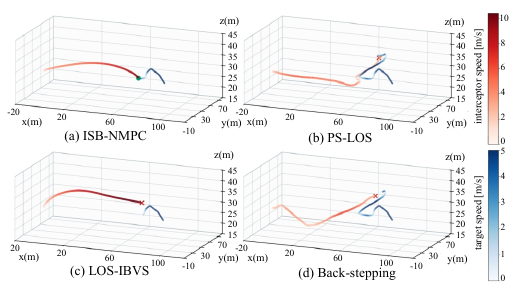}
    \caption{Paired simulation trajectories:
    (a) ISB-NMPC, (b) PS-LOS, (c) LOS-IBVS, and (d) backstepping. All methods
    share target motion and initial geometry. Colors indicate speed;
    green markers denote interception and red crosses denote failure.}
    \label{fig:traj_compare}
\end{figure}

\newpage
\textbf{Interception Algorithm Comparison.} Baselines are planar-sector LOS (PS-LOS)
guidance~\cite{liuPlanarsectorGuidanceInterception}, LOS image-based visual
servoing (LOS-IBVS), and backstepping~\cite{yangHighSpeedInterceptionMulticopter2025a}.
LOS-IBVS combines proportional bearing feedback with forward-speed and yaw-rate
commands~\cite{chaumetteVisualServoControl2006}. All methods share target
states, trajectories, bearings, and initial conditions
(Table~\ref{tab:controller_comparison} and Fig.~\ref{fig:traj_compare}).

Figure~\ref{fig:sim_controller_comparison} summarizes the
ISB-NMPC simulation trajectories.

The joint criteria in Sec.~\ref{sec:visual_interception_problem}
are reported in Table~\ref{tab:controller_comparison}: ISB-NMPC succeeds in
26/30 trials (86.7\%) with 99.11\% mean target-visible time, and is faster
than PS-LOS and backstepping among successful trials. LOS-IBVS closes faster
but succeeds only twice and frequently loses visibility, illustrating why
closing speed alone is insufficient.
For the
paired seed in Fig.~\ref{fig:traj_compare}, ISB-NMPC is the only successful
controller while retaining aggressive closing. Its four failures coincide
with sudden lateral target maneuvers, but the logs do not isolate the effects
of prediction, control response, and active constraints.

\subsection{Experimental Setup and Flight Validation}

\subsubsection{Experimental Platform}

The 1.3~kg, 7-inch interceptor uses an IMX577 camera
($2592\times1944$, 30~Hz, 2.8~mm focal length,
$82^\circ\times66^\circ$ calibrated FOV) and a 50~Hz bearing filter.
PX4 EKF supplies ego-state; no target position or velocity is provided.
Bearing-only prediction uses zero target velocity and nominal range equal to
the configured maximum interception speed times the horizon duration.
A separate quadrotor tows the moving target
(Fig.~\ref{fig:flight_dynamic_validation}).

\subsubsection{Experiment 1: Long-Range Interception of a Quasi-Static Small Target}

Detection progressed from REACQUIRE-HR through TRACK-HR to TRACK-LR as the
apparent target size increased, while interceptor speed reached 17~m/s
(Fig.~\ref{fig:flight_validation}).

\subsubsection{Experiment 2: Pursuit--Evasion Interception of a Dynamic Target}

\begin{figure}[!htbp]
    \centering
    \includegraphics[width=\columnwidth]{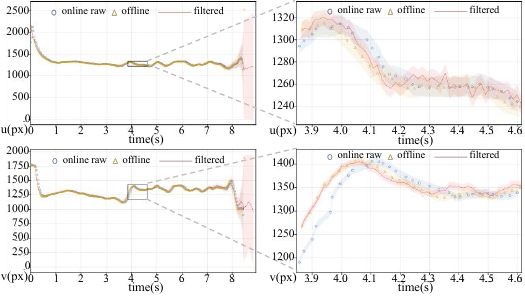}
    \caption{Target image coordinates from online detections,
    offline results, and filtered estimates: $u$ (top) and $v$ (bottom).
    Right panels enlarge the highlighted intervals.}
    \label{fig:flight_filter_comparison}
\end{figure}

The target followed an undulating path at approximately
$5\,\mathrm{m\,s^{-1}}$ (Fig.~\ref{fig:flight_dynamic_validation}).
Figure~\ref{fig:flight_filter_comparison} compares online detections and
DC-$\mathbb S^2$EKF estimates with offline detections at acquisition timestamps.
The filter follows the offline trend, particularly in the vibration-sensitive
$v$ direction, and supplies bearings between detections. This comparison is
qualitative because offline detections are not independent ground truth.

\begin{figure}[!t]
    \centering
    \includegraphics[width=\columnwidth]{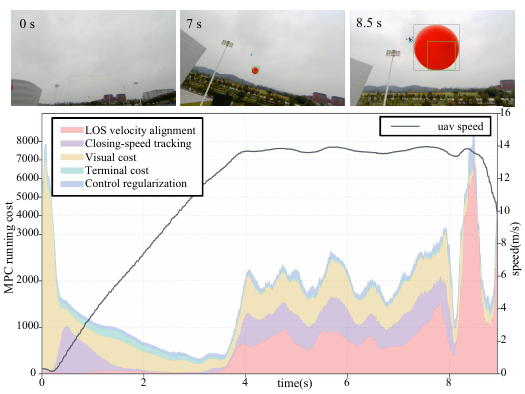}
    \caption{Flight interception: onboard images at 0, 7,
    and 8.5~s (top); stacked MPC cost components and interceptor speed
    (right axis) below.}
    \label{fig:flight_mpc_data}
\end{figure}

Figure~\ref{fig:flight_mpc_data} complements the joint performance
criteria with onboard observations: visual cost dominates near the image
boundary, while LOS velocity-alignment cost grows near encounter.
Speed reaches approximately 14~m/s within 4~s and is largely sustained.
These observations are consistent with the modeled coupling, not a causal
test or proof of constraint satisfaction.

\section{Discussion and Conclusion}

The paired simulations show that closing speed should be coordinated with
visibility rather than maximized alone: ISB-NMPC achieved
\textbf{86.7\% success} and \textbf{99.11\% mean target-visible time}, whereas
LOS-IBVS was faster when successful but succeeded only twice. Jointly
optimizing desired closing speed and vehicle motion therefore provides a
mechanism for allocating shared attitude authority between approach and visual
retention, although the comparison does not isolate individual cost terms.

Simulations with ground-truth target inputs isolate controller behavior,
whereas bearing-only outdoor flights demonstrate the integrated
perception--control pipeline. Long-range acquisition and maneuvering-target
pursuit complement rather than replace the paired comparison; agreement with
offline detections is qualitative, not independent bearing-accuracy evidence.

Flight prediction assumes nominal range and zero target velocity, which can
limit prediction during strong maneuvers. The four simulation failures and
long occlusions motivate better lateral-motion prediction and active-constraint
analysis. Neither recursive feasibility nor capture against arbitrary target
motion is guaranteed, and the typical 10~ms solve time is not a worst-case
bound. Future work should address range sensing, maneuver prediction,
uncertainty-aware reacquisition, and component ablations.

In summary, ISB-NMPC connects LOS-aligned closing objectives to spherical
visual prediction under quadrotor dynamics. The evaluated scenarios support
its utility, while prediction uncertainty and sustained visibility remain key
deployment boundaries.

\begingroup
\renewcommand{\footnotesize}{\fontsize{8}{8.5}\selectfont}
\bibliographystyle{IEEEtran}
\bibliography{tmech-review-base,tmech-review-references}
\endgroup

\begingroup
\raggedbottom
\makeatletter
\def\@IEEEBIOskipN{32pt}
\expandafter\patchcmd\csname\string\IEEEbiography\endcsname
  {\vskip \@IEEEBIOskipN plus 1fil minus 0\baselineskip}
  {\vskip \@IEEEBIOskipN}
  {}{\PackageWarning{paperlayout}{Biography spacer patch failed}}
\BeforeBeginEnvironment{IEEEbiography}{%
  \ifnum\value{IEEEbiography}=4\relax\newpage\fi}
\makeatother

\begin{IEEEbiography}[{\includegraphics[width=1in,height=1.25in,trim=0 11.5bp 0 11.5bp,clip]{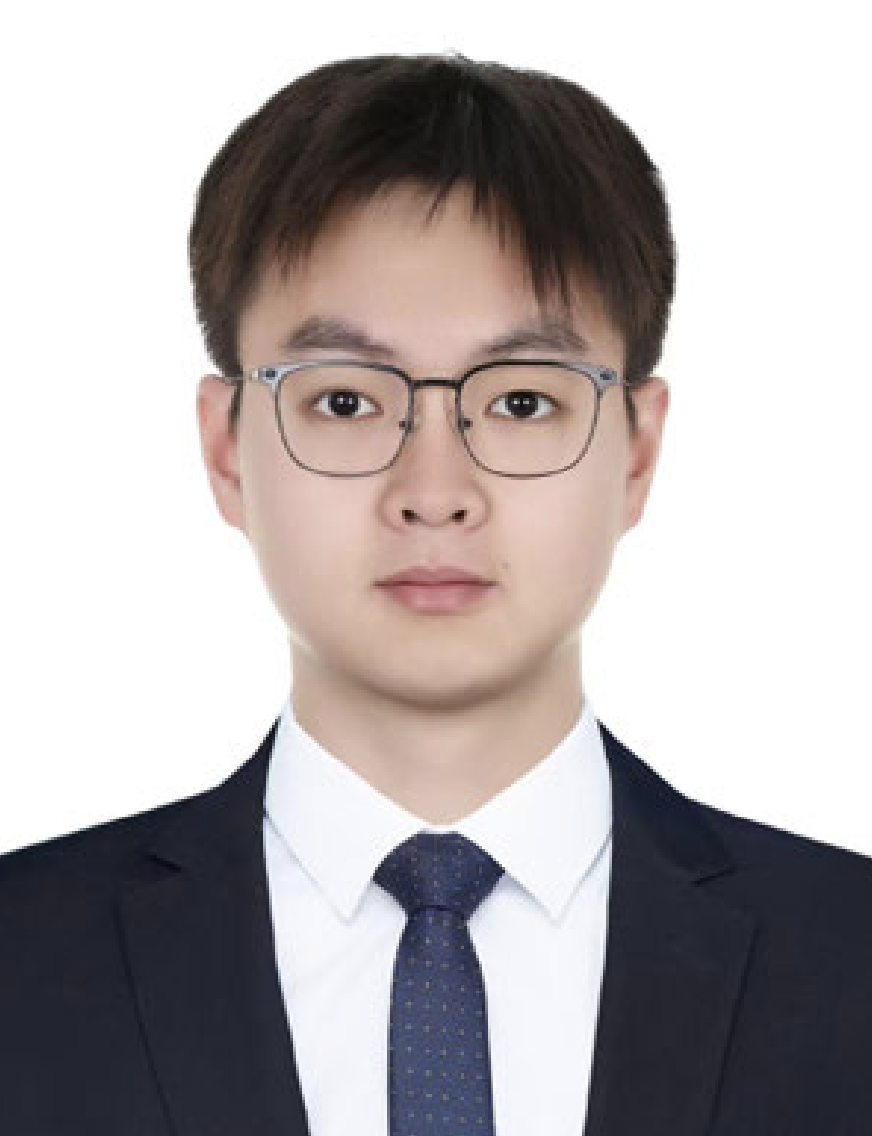}}]{Li Hong}
received the B.S. degree in control science and engineering from Shen Yuan Honors College, Beihang University, Beijing, China, in 2026, where he is currently pursuing the Ph.D. degree in control science and engineering. His main research interests include UAV navigation, planning, and control.
\end{IEEEbiography}

\begin{IEEEbiography}[{\includegraphics[width=1in,height=1.25in,trim=0 26.25bp 0 26.25bp,clip]{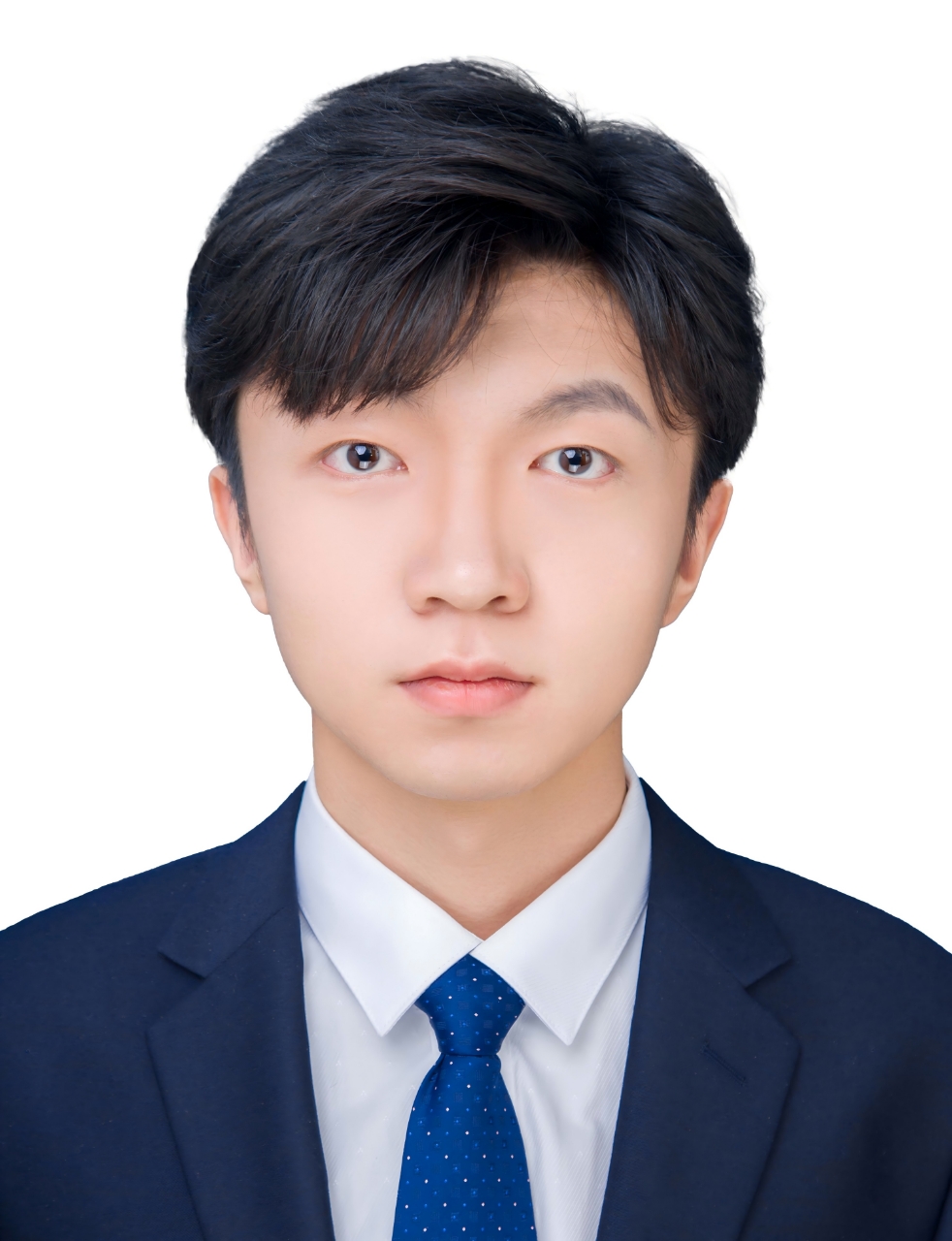}}]{Linkai Liu}
is currently pursuing the Ph.D. degree in Navigation, Guidance and Control at the School of Automation Science and Electrical Engineering, Beihang University (formerly Beijing University of Aeronautics and Astronautics), Beijing, China. He received the B.S. degree in Intelligent Flight Vehicle Technology from Beihang University in 2025. His main research interests include vision-based guidance and control, and autonomous aerial robotics.
\end{IEEEbiography}

\begin{IEEEbiography}[{\includegraphics[width=1in,height=1.25in,trim=0 9.57bp 0 9.57bp,clip]{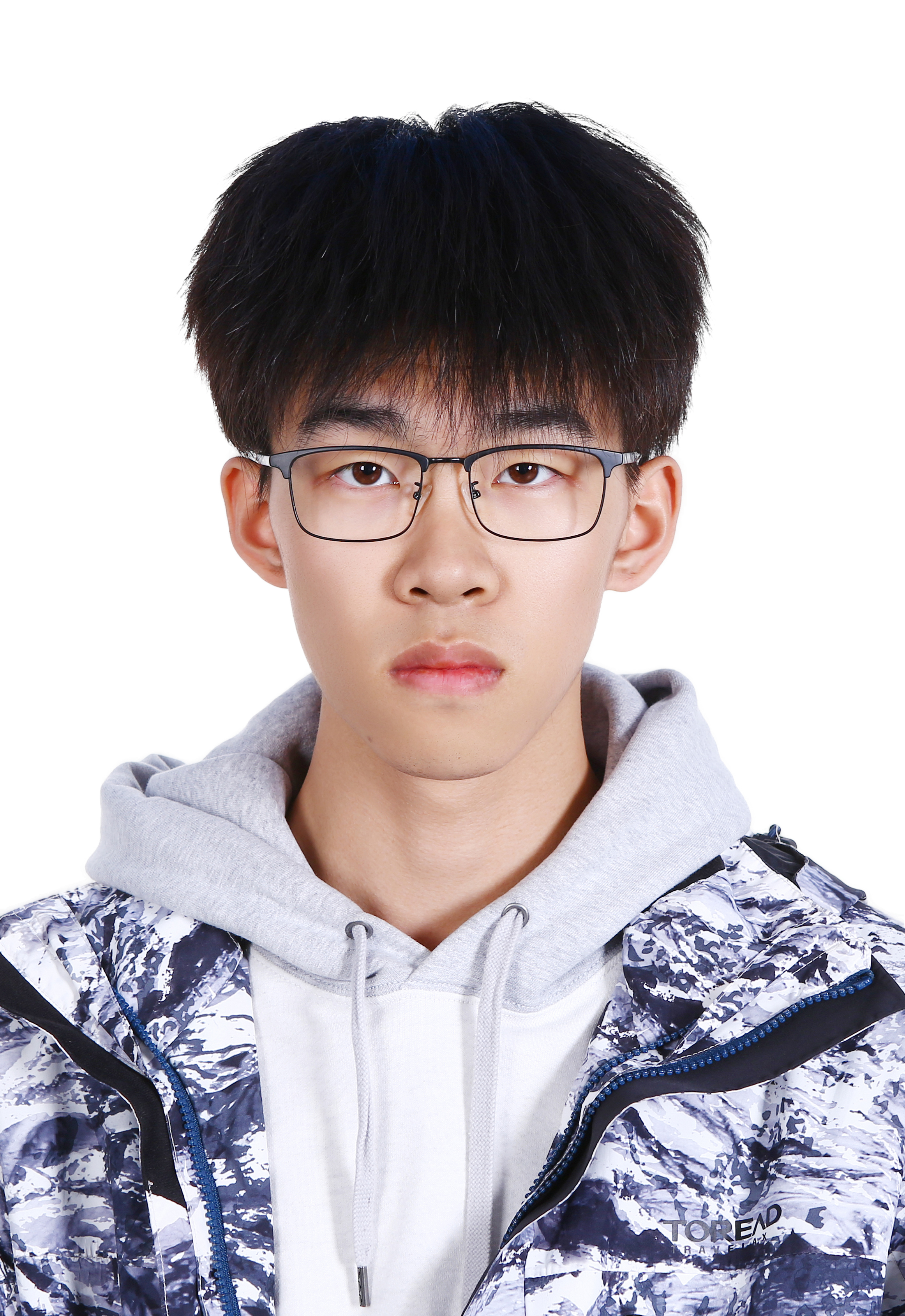}}]{Xiangyu Ren}
is currently pursuing the master's degree in control engineering with the School of Automation Science and Electrical Engineering, Beihang University, Beijing, China. His research interests include aerial robotics.
\end{IEEEbiography}

\begin{IEEEbiography}[{\includegraphics[width=1in,height=1.25in,trim=0 1.75bp 0 1.75bp,clip]{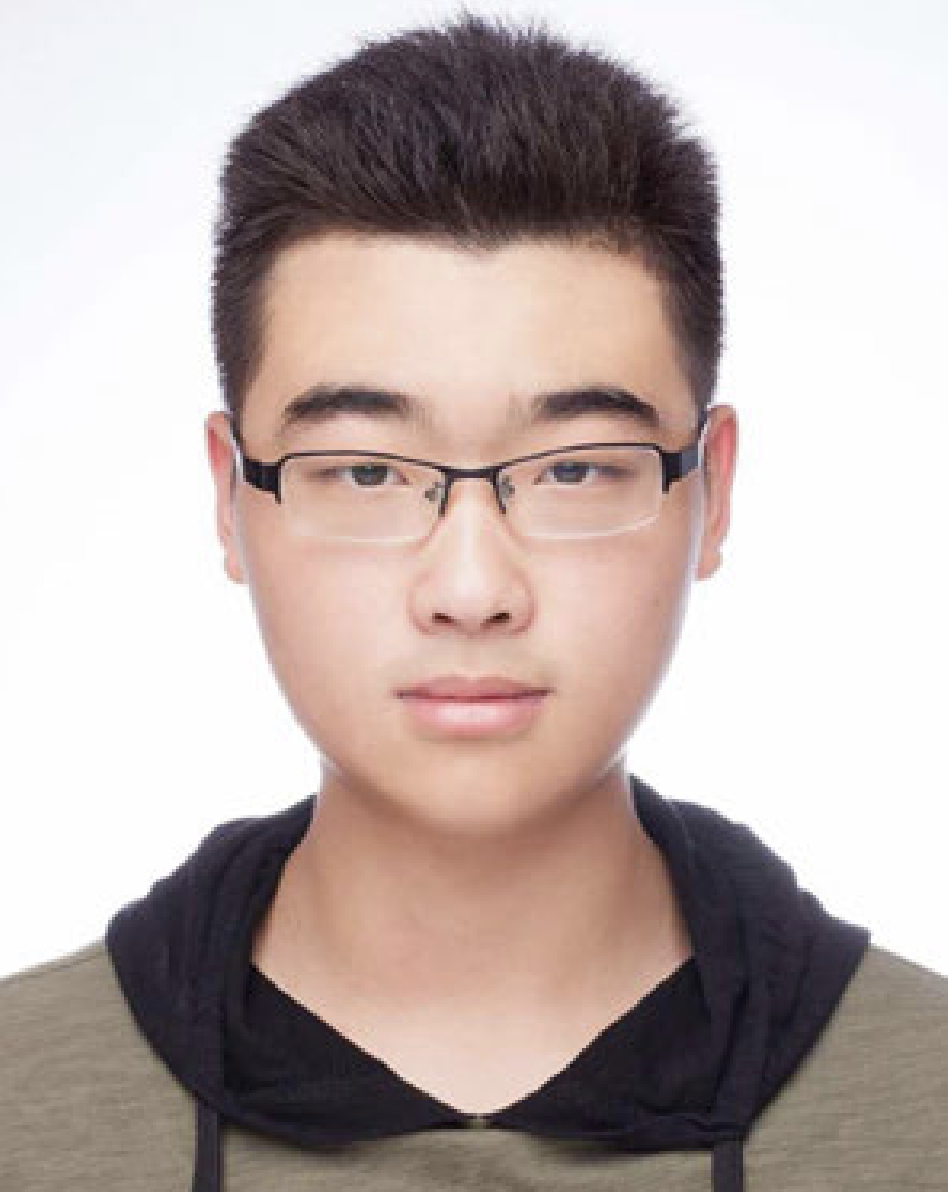}}]{Shuli Lv}
received the B.S. degree in control science and engineering from Shen Yuan Honors College, Beihang University, Beijing, China, in 2022. He is currently pursuing the Ph.D. degree in control science and engineering with the School of Automation Science and Electrical Engineering, Beihang University. His main research interests include UAV navigation, planning, and control.
\end{IEEEbiography}

\begin{IEEEbiography}[{\includegraphics[width=1in,height=1.25in,trim=0 99bp 0 99bp,clip]{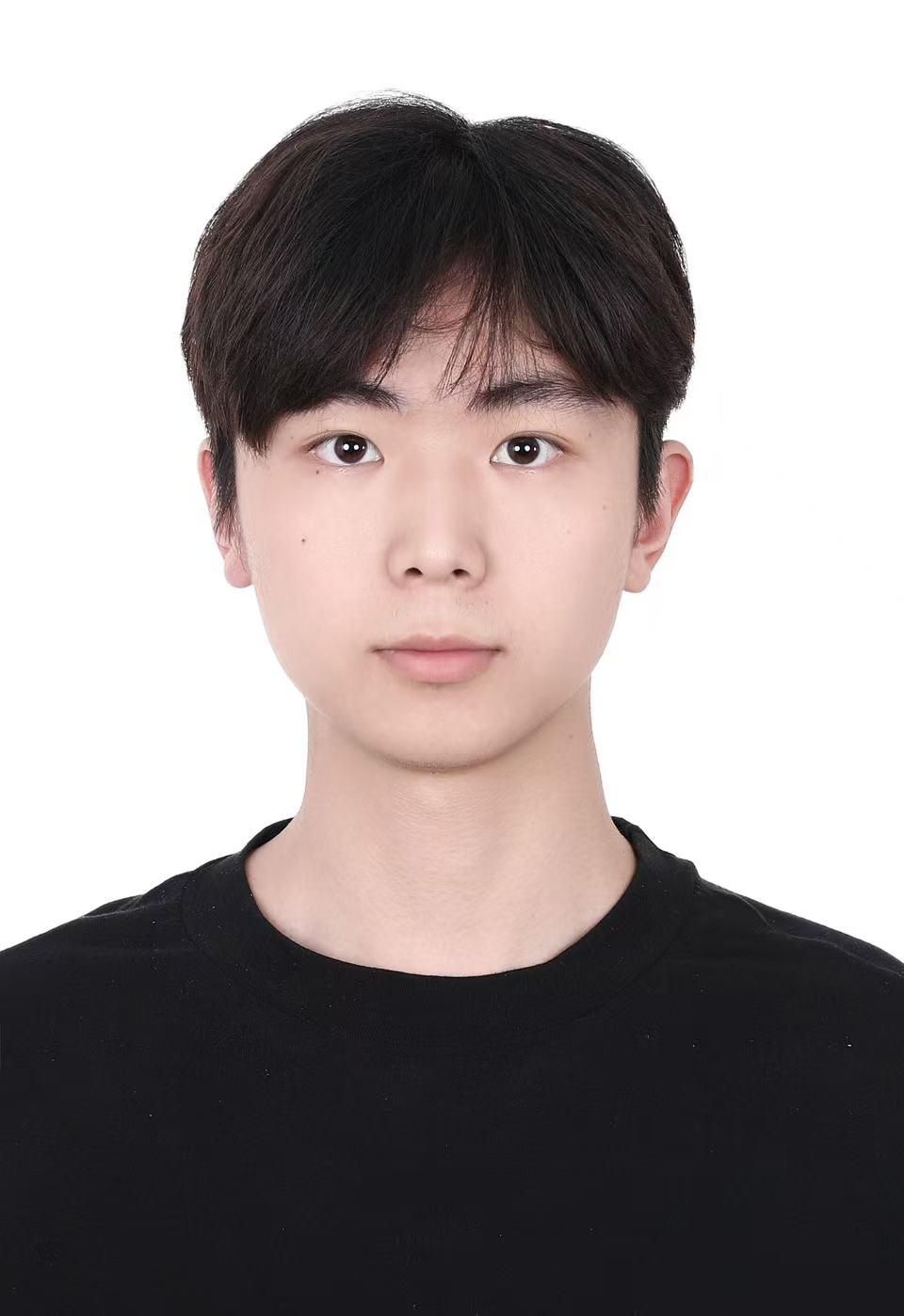}}]{Chen Min}
received the B.S. degree in control science and engineering from Shen Yuan Honors College, Beihang University, Beijing, China, in 2025. He is currently working toward the Ph.D. degree in control science and engineering with Shen Yuan Honors College, Beihang University, Beijing, China. His main research interests include UAV navigation, planning, and control.
\end{IEEEbiography}

\begin{IEEEbiography}[{\includegraphics[width=1in,height=1.25in,trim=0 96.125bp 0 96.125bp,clip]{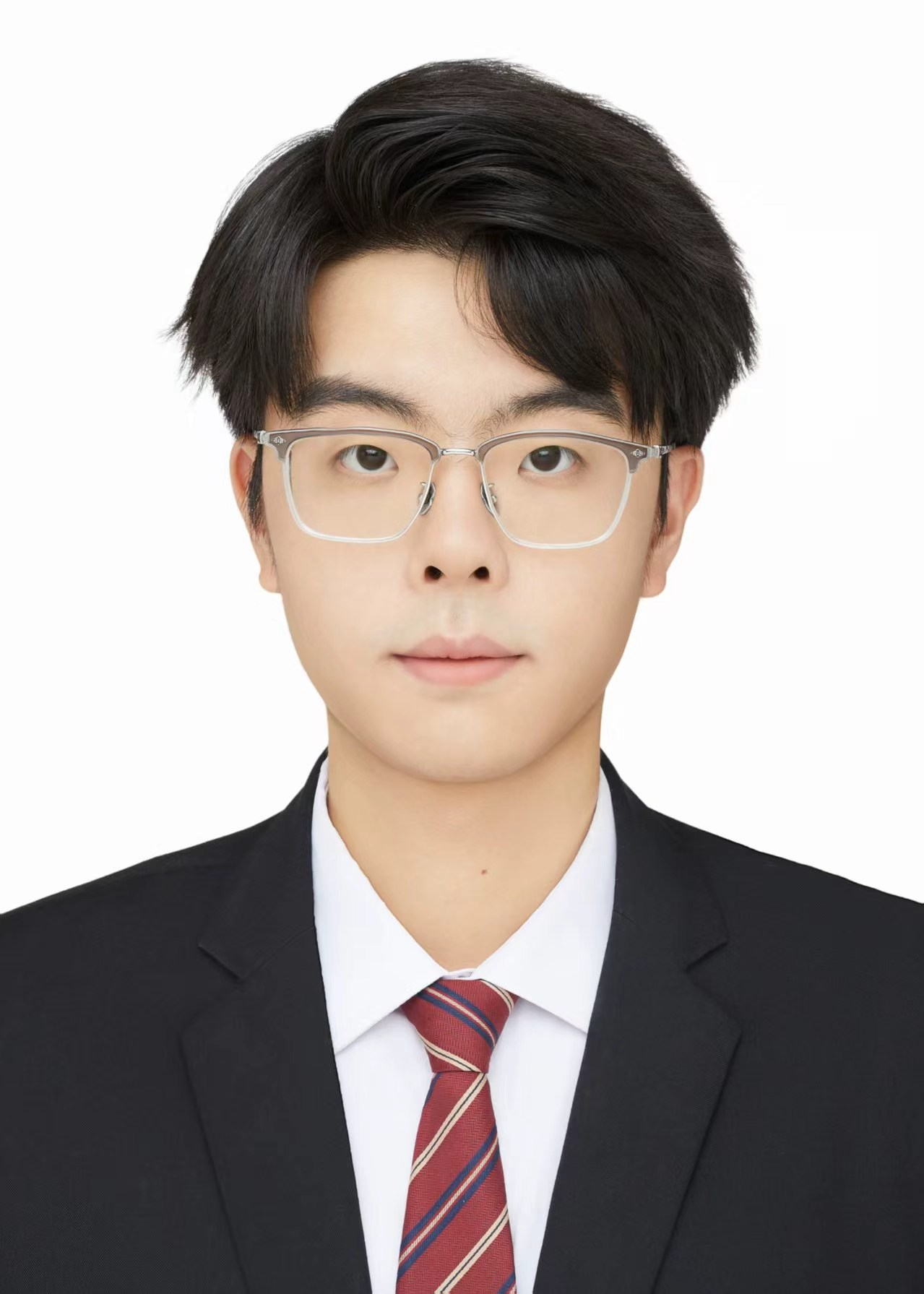}}]{Chunjiang Wu}
received the B.S. degree in automation from North China Electric Power University, China, in 2023, and the M.S. degree in control science and engineering from the School of Automation Science and Electrical Engineering, Beihang University, Beijing, China, in 2026. He is currently pursuing the Ph.D. degree in low-altitude intelligent navigation technology at the Hangzhou International Innovation Institute of Beihang University, Hangzhou, China. His research interests include visual recognition and embodied intelligence.
\end{IEEEbiography}

\begin{IEEEbiography}[{\includegraphics[width=1in,height=1.25in,trim=0 53.625bp 0 53.625bp,clip]{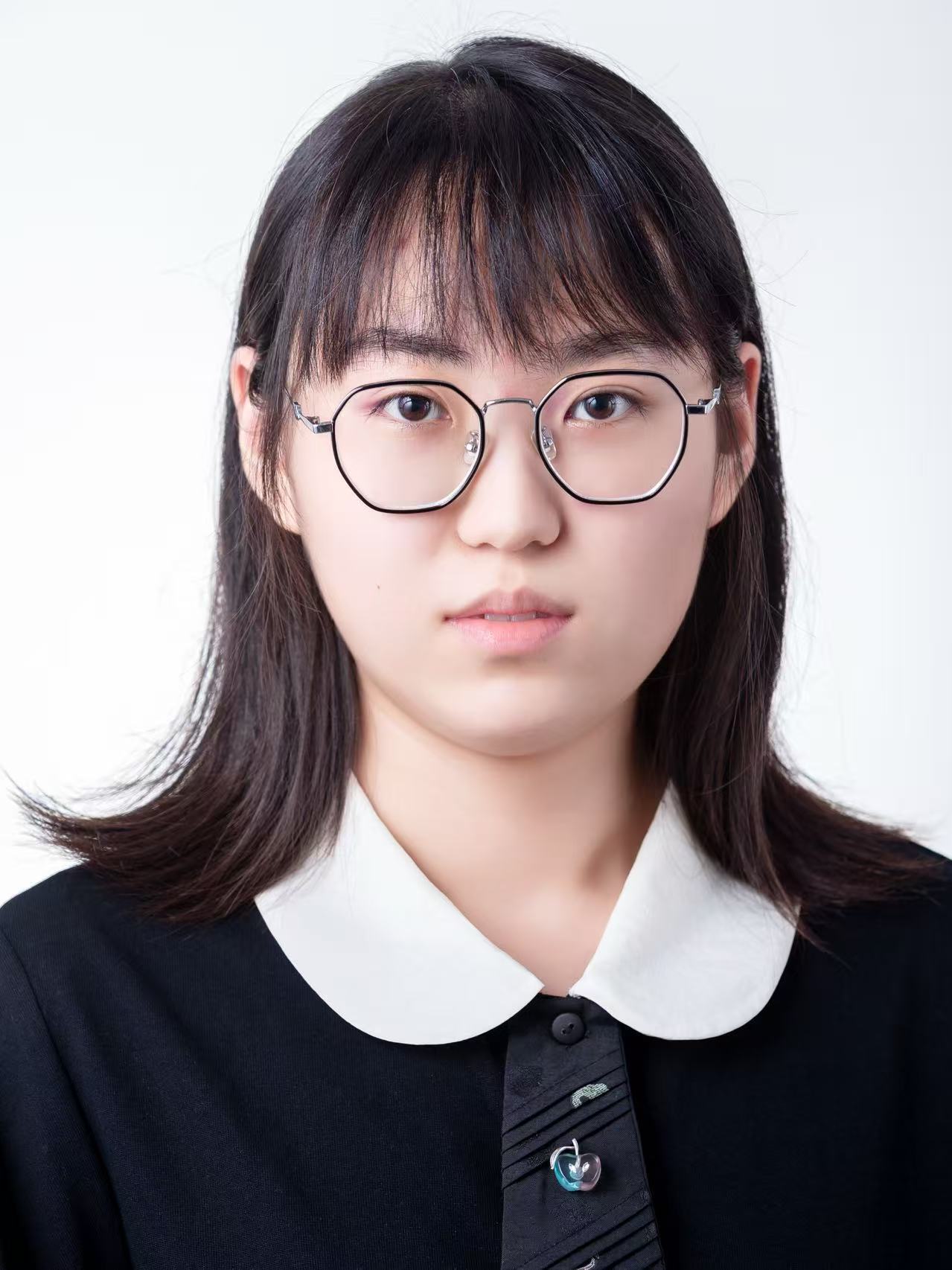}}]{Yujun Zhang}
is currently pursuing the Ph.D. degree in Control Science and Engineering at Shen Yuan Honors College, Beihang University, Beijing, China. She received the B.S. degree in Automation from Beihang University in 2026. Her research interests include pursuit-evasion games, multi-UAV cooperative planning, and autonomous aerial robotics.
\end{IEEEbiography}

\begin{IEEEbiography}[{\includegraphics[width=1in,height=1.25in,trim=0 60bp 0 60bp,clip]{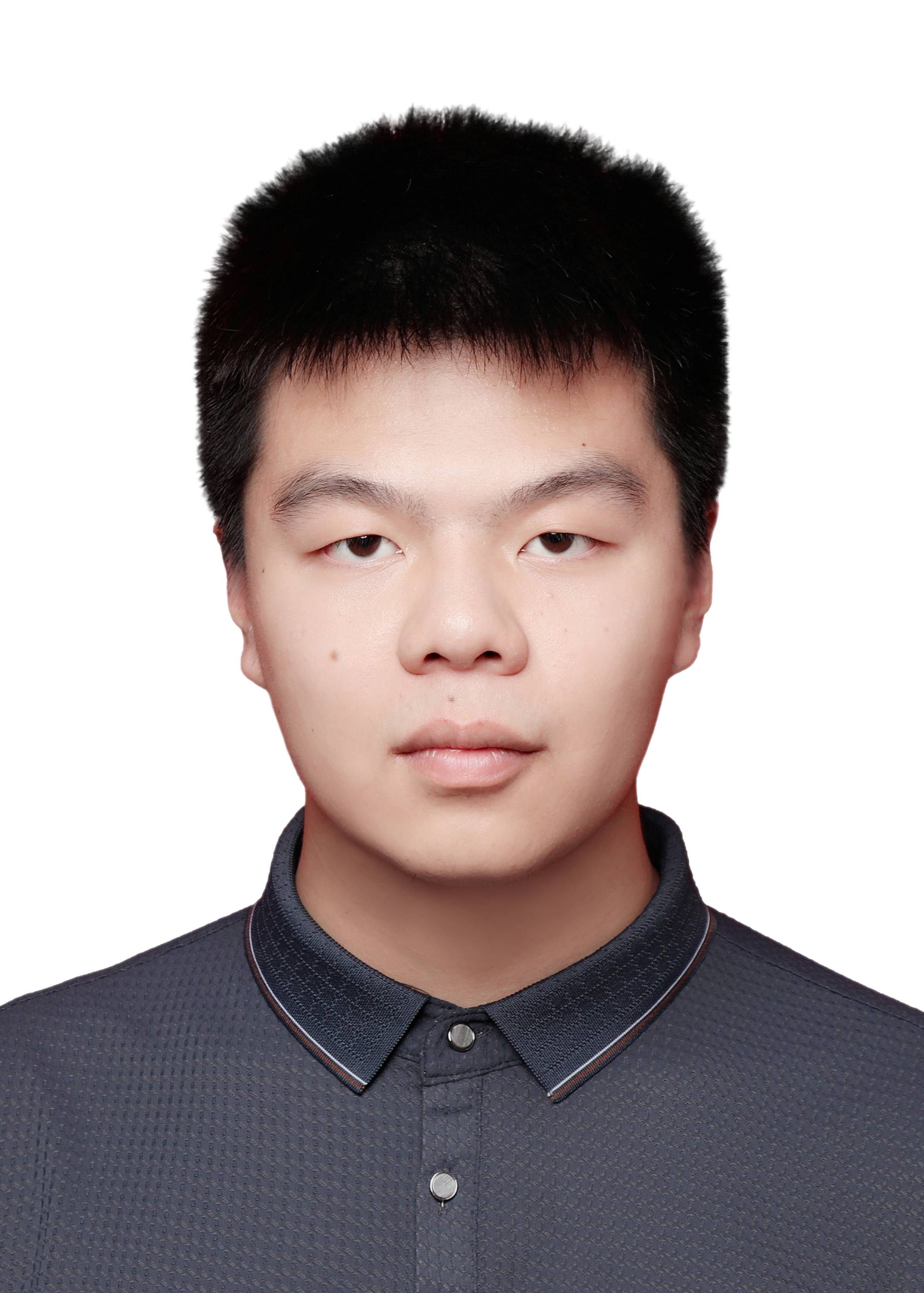}}]{Dingchen Zeng}
is currently pursuing a bachelor's degree in automation at the School of Automation Science and Electrical Engineering, Beihang University, Beijing, China. Current research interests include autonomous aerial robotics and embodied intelligence.
\end{IEEEbiography}

\begin{IEEEbiography}[{\includegraphics[width=1in,height=1.25in,trim=0 13.75bp 0 13.75bp,clip]{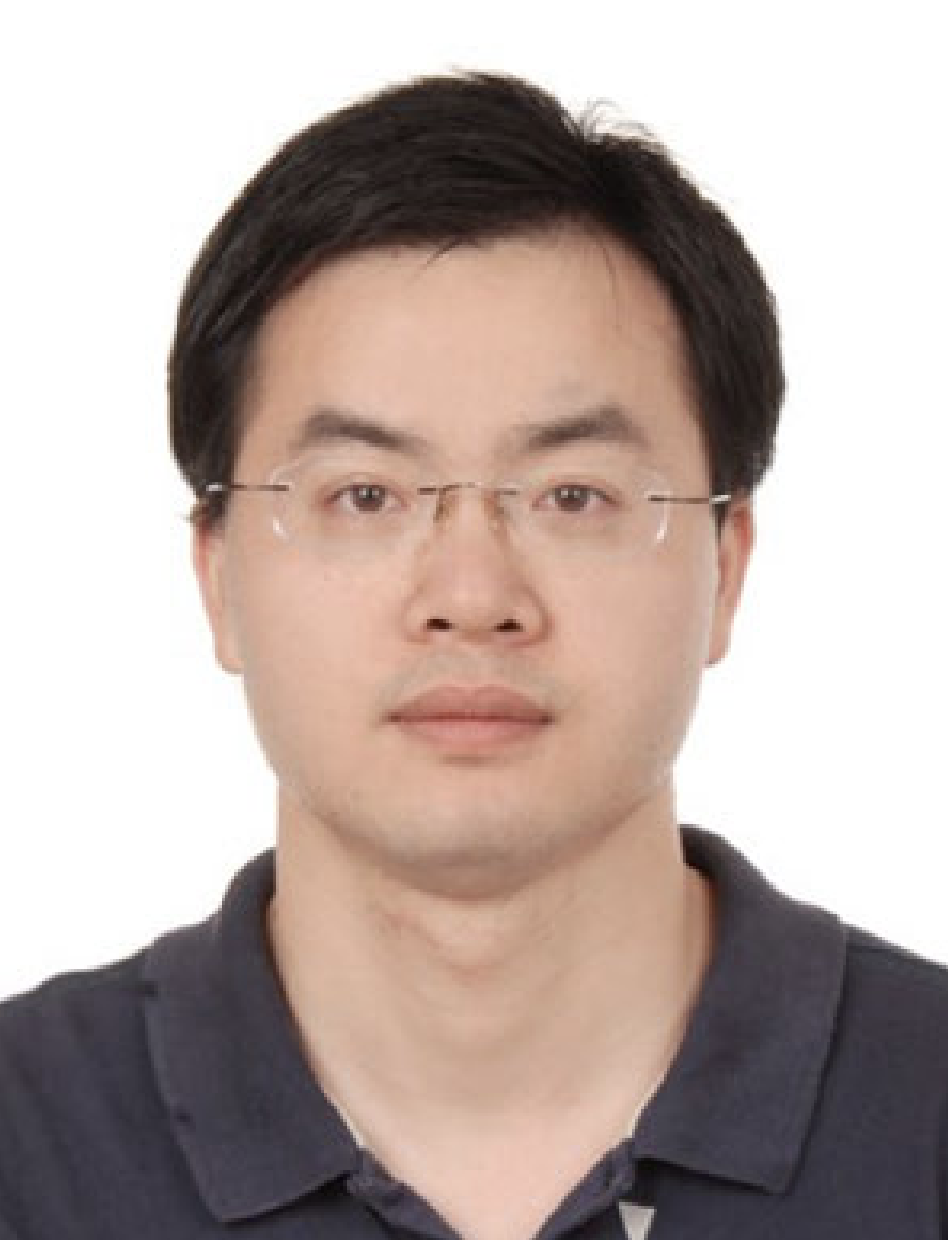}}]{Quan Quan}
(Senior Member, IEEE) received the B.S. and Ph.D. degrees in control science and engineering from Beihang University, Beijing, China, in 2004 and 2010, respectively. Since 2022, he has been a Professor with Beihang University, where he is currently with the School of Automation Science and Electrical Engineering. His research interests include reliable flight control, swarm intelligence, vision-based navigation, and health assessment.
\end{IEEEbiography}

\endgroup

\end{document}